\documentclass[runningheads]{llncs}
\pdfoutput=1
 
\usepackage{eccv}

\usepackage{eccvabbrv}

\usepackage{graphicx}
\usepackage{booktabs}
\usepackage{makecell}
\usepackage{pifont}
\usepackage{tikz}
\usepackage{siunitx}

\usepackage{mathtools}
\usepackage{multirow}
\usepackage{array}
\usepackage{pgfplots}
\usepackage{adjustbox}

\usepackage{microtype}
\usepackage[table]{xcolor}
\pgfplotsset{compat=1.18}
\usepackage[figuresright]{rotating}

\usepackage{algorithm}
\usepackage{algpseudocode}

\newcommand{\ours}{RAS\xspace}

\definecolor{light-gray}{gray}{0.7}

\newcommand{\tabcompare}{%
\newcommand{\cmark}{\large {\textcolor{green!60!black}{\ding{51}}}}
\newcommand{\xmark}{\large {\textcolor{red}{\ding{55}}}}
\begin{table*}[t]
\centering
\caption{\textit{Technical Comparison of OoD Detection Methods}, extended from~\cite{liu2023gen}.
The table summarizes the requirements and capabilities of established OOD detection methods, and highlights the additional capabilities provided by \ours, namely requiring only unlabelled ID activations to compute $\boldsymbol{\mu}$, introducing no tunable threshold at inference, and operating on both features and logits. Notably, \ours makes no assumption about the sign of the penultimate layer activations, making it applicable to architectures such as ViTs~\cite{dosovitskiy2020vit} and ConvNeXt~\cite{liu2022convnet}, unlike ASH-S and SCALE, which tend to fail when negative activations skew the $Q / Q_p$ ratio.
}
\label{tab:compare}
\renewcommand{\arraystretch}{1.45}
\setlength{\tabcolsep}{5pt}
\begin{adjustbox}{max width=\linewidth}
\begin{tabular}{l l p{7.0cm} ccc cc c}
\toprule
& \multirow{2}{*}{\textbf{Method}}
  & \multirow{2}{*}{\textbf{Equation}}
  & \multicolumn{3}{c}{\textbf{Free of}}
  & \multicolumn{2}{c}{\textbf{Leverages}}
  & \multirow{2}{*}{\shortstack{\textbf{Applicable to}\\\textbf{Transformers}}}
  \\
\cmidrule(lr){4-6}\cmidrule(lr){7-8}
& &
  & ID access & ID labels & hyperparam.
  & features & logits
  &
  \\
\midrule
\multirow{4}{*}{\rotatebox[origin=c]{90}{\scriptsize\textit{Scoring}}}
& EBO~\cite{liu2020energy}
  & $\log \sum_k e^{f_k(\mathbf{a})}$
  & \cmark & \cmark & \cmark
  & & \cmark
  & \cmark \\
& GEN~\cite{liu2023gen}
  & $G_\gamma(\mathbf{p}) = -\sum_{m=1}^{M} p_{i_m}^{\gamma}(1-p_{i_m})^{\gamma},\; p_{i_1}\geq\cdots\geq p_{i_C}$
  & \cmark & \cmark & \xmark\;{\small$(\gamma)$}
  & & \cmark
  & \cmark \\
& ViM~\cite{haoqi2022vim}
  & $-\alpha\|\mathbf{a}^{P^\perp}\|_2 + \operatorname{LogSumExp} f(\mathbf{a})$
  & \xmark\;{\small$(\alpha, P)$} & \cmark & \xmark\;{\small$(\alpha)$}
  & \cmark & \cmark
  & \cmark \\
& Maha~\cite{lee2018simple}
  & $\max_c -(\mathbf{a} - \hat{\boldsymbol{\mu}}_c)^\top \hat{\boldsymbol{\Sigma}}^{-1} (\mathbf{a} - \hat{\boldsymbol{\mu}}_c)$
  & \xmark\;{\small$(\hat{\boldsymbol{\Sigma}}, \hat{\boldsymbol{\mu}}_c)$} & \xmark & \cmark
  & \cmark &
  & \cmark \\
\midrule
\multirow{7}{*}{\rotatebox[origin=c]{90}{\scriptsize\textit{Enhancing}}}
& ReAct~\cite{sun2021react}
  & $\tilde{\mathbf{a}} = \min(\mathbf{a},\, b)$
  & \xmark\;{\small$(b)$} & \cmark & \xmark\;{\small$(b)$}
  & \cmark & \cmark
  & \cmark \\
& ASH-P~\cite{djurisic2023extremely}
  & $\tilde{a}_j = a_j \cdot \mathbf{1}[a_j > P_p(\mathbf{a})]$
  & \cmark & \cmark & \xmark\;{\small$(p)$}
  & \cmark & \cmark
  & \cmark \\
& ASH-B~\cite{djurisic2023extremely}
  & $\tilde{a}_j = \bar{a}_p \cdot \mathbf{1}[a_j > P_p(\mathbf{a})]$,\;
    $\bar{a}_p = \operatorname{mean}_{a_j > P_p}(a_j)$
  & \cmark & \cmark & \xmark\;{\small$(p)$}
  & \cmark & \cmark
  & \xmark \\
& ASH-S~\cite{djurisic2023extremely}
  & $\tilde{a}_j = a_j\exp(r)\cdot\mathbf{1}[a_j > P_p(\mathbf{a})]$,\;
    $r = Q/Q_p$
  & \cmark & \cmark & \xmark\;{\small$(p)$}
  & \cmark & \cmark
  & \xmark \\
& SCALE~\cite{xu2024scaling}
  & $\tilde{a}_j = a_j\exp(r)\ \forall j$,\;
    $r = {\sum_j a_j}\,/\!\!\sum_{a_j > P_p}\!\!a_j$
  & \cmark & \cmark & \xmark\;{\small$(p)$}
  & \cmark & \cmark
  & \xmark \\
& DICE~\cite{sun2022dice}
  & $\tilde{\mathbf{W}} = \mathbf{W} \odot \mathbf{M}_s$;\quad top-$s$ weight-contribution mask
  & \xmark\;{\small$(s)$} & \cmark & \xmark\;{\small$(s)$}
  & & \cmark
  & \cmark \\
\rowcolor{gray!20}
& \textbf{\ours (ours)}
  & $\bar{\mathbf{a}}_{\pi(j)} \leftarrow \mu_j$; 
    $\boldsymbol{\mu} = \tfrac{1}{N}\!\sum_i \operatorname{r}(\mathbf{a}_i)$, 
    $\pi\!: a_{\pi(1)}\!\geq\!\cdots\!\geq\! a_{\pi(d)}$
  & \xmark\;{\small$(\boldsymbol{\mu})$} & \cmark
  & \cmark & \cmark
  & \cmark & \cmark \\
\bottomrule
& \multicolumn{8}{p{20cm}}{\emph{Notation:} penultimate-layer activations $\mathbf{a}$, logits $f(\mathbf{a})$, prediction distribution $\mathbf{p} = \operatorname{softmax}(f(\mathbf{a}))$, $p$-th percentile $P_p(\mathbf{a})$, null space of the top principal directions of ID activations $P^\perp$, $Q = \sum_j a_j$, $Q_p = \sum_{a_j>P_p} a_j$, per-class means and tied covariance of ID activations $\hat{\boldsymbol{\mu}}_c$ and $\hat{\boldsymbol{\Sigma}}$, mean ranked activations over ID training set $\boldsymbol{\mu}$, rank-order permutation of test activations $\pi$, and ranking function $\operatorname{r}(\cdot)$.}
\end{tabular}
\end{adjustbox}
\end{table*}%
}

\newcommand{\tabinhibitexcite}{%
\begin{table}[t!]
\centering
\caption{Ablation of \ours into \ours-inhibit (activations above $\boldsymbol{\mu}$ shifted down) and \ours-excite (activations below $\boldsymbol{\mu}$ shifted up). EBO ($\mu{\pm}\sigma$) is the baseline score distribution for each OoD dataset. $\Delta\mu$ and $\Delta$AUROC are relative to EBO scores. ID is shown in bold in each header.}
\resizebox{\linewidth}{!}{
\begin{tabular}{l c  ccc  ccc  ccc}
\toprule
& \textbf{EBO} & \multicolumn{3}{c}{\textbf{EBO + \ours -excite}} & \multicolumn{3}{c}{\textbf{EBO + \ours -inhibit}} & \multicolumn{3}{c}{\textbf{EBO + \ours}} \\
\cmidrule(lr){2-2}\cmidrule(lr){3-5}\cmidrule(lr){6-8}\cmidrule(lr){9-11}
& $\mu{\pm}\sigma$ & $\mu{\pm}\sigma$ & $\Delta\mu$ & $\Delta$AUC & $\mu{\pm}\sigma$ & $\Delta\mu$ & $\Delta$AUC & $\mu{\pm}\sigma$ & $\Delta\mu$ & $\Delta$AUC \\
\midrule
\textbf{CIFAR-10}  & $9.09{\scriptstyle\,\pm\,1.51}$ & $9.32{\scriptstyle\pm1.24}$ & $+0.23$ & $-$ & $8.47{\scriptstyle\pm1.05}$ & $-0.63$ & $-$ & $8.67{\scriptstyle\pm0.84}$ & $-0.42$ & $-$ \\
\midrule
CIFAR-100 & $6.37{\scriptstyle\pm1.83}$ & $7.04{\scriptstyle\pm1.56}$ & $+0.67$ & $+0.53$ & $6.25{\scriptstyle\pm1.50}$ & $-0.12$ & $+1.82$ & $6.89{\scriptstyle\pm1.32}$ & $+0.52$ & $+0.99$ \\
TIN & $6.07{\scriptstyle\pm1.72}$ & $6.75{\scriptstyle\pm1.53}$ & $+0.68$ & $+0.25$ & $5.95{\scriptstyle\pm1.45}$ & $-0.12$ & $+1.53$ & $6.72{\scriptstyle\pm1.29}$ & $+0.64$ & $+0.00$ \\
MNIST & $5.37{\scriptstyle\pm1.41}$ & $6.34{\scriptstyle\pm1.32}$ & $+0.97$ & $-0.64$ & $5.41{\scriptstyle\pm1.32}$ & $+0.04$ & $+0.03$ & $6.39{\scriptstyle\pm1.26}$ & $+1.02$ & $-1.95$ \\
SVHN & $5.81{\scriptstyle\pm1.54}$ & $6.80{\scriptstyle\pm1.36}$ & $+0.99$ & $-1.52$ & $5.83{\scriptstyle\pm1.37}$ & $+0.02$ & $+0.19$ & $6.81{\scriptstyle\pm1.22}$ & $+1.00$ & $-2.97$ \\
Texture & $6.15{\scriptstyle\pm1.87}$ & $6.80{\scriptstyle\pm1.57}$ & $+0.64$ & $+1.18$ & $5.90{\scriptstyle\pm1.40}$ & $-0.26$ & $+3.60$ & $6.63{\scriptstyle\pm1.26}$ & $+0.47$ & $+2.73$ \\
Places365 & $6.01{\scriptstyle\pm1.77}$ & $6.68{\scriptstyle\pm1.52}$ & $+0.68$ & $+0.98$ & $5.90{\scriptstyle\pm1.46}$ & $-0.10$ & $+2.03$ & $6.55{\scriptstyle\pm1.26}$ & $+0.54$ & $+2.42$ \\
\rowcolor{gray!20}\textit{Average} & $5.96{\scriptstyle\pm1.69}$ & $6.74{\scriptstyle\pm1.47}$ & $+0.77$ & $+0.13$ & $5.87{\scriptstyle\pm1.42}$ & $-0.09$ & $+1.53$ & $6.66{\scriptstyle\pm1.27}$ & $+0.70$ & $+0.20$ \\
\midrule
\textbf{CIFAR-100} & $11.14{\scriptstyle\,\pm\,3.75}$ & $11.71{\scriptstyle\pm3.13}$ & $+0.57$ & $-$ & $10.01{\scriptstyle\pm2.35}$ & $-1.13$ & $-$ & $10.65{\scriptstyle\pm1.84}$ & $-0.49$ & $-$ \\
\midrule
CIFAR-10 & $7.74{\scriptstyle\pm2.07}$ & $8.95{\scriptstyle\pm1.80}$ & $+1.21$ & $-0.83$ & $7.59{\scriptstyle\pm1.69}$ & $-0.14$ & $+0.26$ & $8.84{\scriptstyle\pm1.51}$ & $+1.11$ & $-1.52$ \\
TIN & $7.35{\scriptstyle\pm1.79}$ & $8.46{\scriptstyle\pm1.64}$ & $+1.11$ & $+0.52$ & $7.18{\scriptstyle\pm1.47}$ & $-0.17$ & $+1.30$ & $8.32{\scriptstyle\pm1.35}$ & $+0.96$ & $+1.26$ \\
MNIST & $7.68{\scriptstyle\pm1.67}$ & $8.89{\scriptstyle\pm1.50}$ & $+1.21$ & $-0.26$ & $7.56{\scriptstyle\pm1.49}$ & $-0.12$ & $+0.48$ & $8.73{\scriptstyle\pm1.37}$ & $+1.05$ & $-0.07$ \\
SVHN & $7.39{\scriptstyle\pm1.79}$ & $8.43{\scriptstyle\pm1.56}$ & $+1.04$ & $+1.31$ & $7.20{\scriptstyle\pm1.45}$ & $-0.20$ & $+1.56$ & $8.27{\scriptstyle\pm1.29}$ & $+0.88$ & $+2.27$ \\
Texture & $7.80{\scriptstyle\pm2.14}$ & $8.57{\scriptstyle\pm1.82}$ & $+0.77$ & $+3.47$ & $7.41{\scriptstyle\pm1.54}$ & $-0.39$ & $+2.93$ & $8.27{\scriptstyle\pm1.30}$ & $+0.47$ & $+6.10$ \\
Places365 & $7.64{\scriptstyle\pm1.92}$ & $8.69{\scriptstyle\pm1.65}$ & $+1.04$ & $+1.08$ & $7.51{\scriptstyle\pm1.57}$ & $-0.13$ & $+0.42$ & $8.57{\scriptstyle\pm1.40}$ & $+0.93$ & $+1.17$ \\
\rowcolor{gray!20}\textit{Average} & $7.60{\scriptstyle\pm1.90}$ & $8.66{\scriptstyle\pm1.66}$ & $+1.06$ & $+0.88$ & $7.41{\scriptstyle\pm1.54}$ & $-0.19$ & $+1.16$ & $8.50{\scriptstyle\pm1.37}$ & $+0.90$ & $+1.53$ \\
\midrule
\textbf{ImageNet-200} & $13.99{\scriptstyle\,\pm\,4.57}$ & $14.96{\scriptstyle\pm3.81}$ & $+0.97$ & $-$ & $12.37{\scriptstyle\pm2.62}$ & $-1.62$ & $-$ & $13.34{\scriptstyle\pm1.98}$ & $-0.65$ & $-$ \\
\midrule
SSB-Hard & $9.67{\scriptstyle\pm3.14}$ & $11.28{\scriptstyle\pm2.64}$ & $+1.60$ & $+0.77$ & $9.33{\scriptstyle\pm2.35}$ & $-0.34$ & $+0.55$ & $10.94{\scriptstyle\pm2.02}$ & $+1.26$ & $+0.50$ \\
NINCO & $8.88{\scriptstyle\pm2.34}$ & $10.42{\scriptstyle\pm2.05}$ & $+1.54$ & $+1.85$ & $8.77{\scriptstyle\pm1.98}$ & $-0.11$ & $+0.13$ & $10.22{\scriptstyle\pm1.71}$ & $+1.34$ & $+2.29$ \\
Texture & $7.92{\scriptstyle\pm2.33}$ & $9.39{\scriptstyle\pm2.10}$ & $+1.47$ & $+1.65$ & $7.60{\scriptstyle\pm1.69}$ & $-0.32$ & $+1.94$ & $9.13{\scriptstyle\pm1.55}$ & $+1.20$ & $+3.04$ \\
iNaturalist & $7.80{\scriptstyle\pm1.49}$ & $8.94{\scriptstyle\pm1.37}$ & $+1.14$ & $+3.19$ & $7.65{\scriptstyle\pm1.20}$ & $-0.15$ & $+1.30$ & $8.80{\scriptstyle\pm1.18}$ & $+1.00$ & $+3.75$ \\
Openimage-O & $8.25{\scriptstyle\pm1.85}$ & $9.80{\scriptstyle\pm1.69}$ & $+1.55$ & $+1.84$ & $8.10{\scriptstyle\pm1.55}$ & $-0.15$ & $+1.12$ & $9.59{\scriptstyle\pm1.46}$ & $+1.33$ & $+2.66$ \\
\rowcolor{gray!20}\textit{Average} & $8.51{\scriptstyle\pm2.23}$ & $9.97{\scriptstyle\pm1.97}$ & $+1.46$ & $+1.86$ & $8.29{\scriptstyle\pm1.75}$ & $-0.22$ & $+1.01$ & $9.73{\scriptstyle\pm1.58}$ & $+1.23$ & $+2.45$ \\
\midrule
\textbf{ImageNet} & $17.16{\scriptstyle\,\pm\,4.38}$ & $17.94{\scriptstyle\pm4.03}$ & $+0.78$ & $-$ & $16.16{\scriptstyle\pm3.29}$ & $-0.99$ & $-$ & $16.94{\scriptstyle\pm3.12}$ & $-0.21$ & $-$ \\
\midrule
SSB-Hard & $14.02{\scriptstyle\pm3.25}$ & $15.07{\scriptstyle\pm2.98}$ & $+1.05$ & $-0.38$ & $13.56{\scriptstyle\pm2.74}$ & $-0.46$ & $+0.64$ & $14.61{\scriptstyle\pm2.71}$ & $+0.60$ & $-0.80$ \\
NINCO & $13.14{\scriptstyle\pm2.55}$ & $13.70{\scriptstyle\pm2.46}$ & $+0.56$ & $+3.37$ & $12.69{\scriptstyle\pm2.24}$ & $-0.45$ & $+1.61$ & $13.32{\scriptstyle\pm2.17}$ & $+0.18$ & $+3.45$ \\
iNaturalist & $11.39{\scriptstyle\pm1.77}$ & $11.76{\scriptstyle\pm1.51}$ & $+0.37$ & $+3.55$ & $10.81{\scriptstyle\pm1.30}$ & $-0.58$ & $+3.12$ & $11.24{\scriptstyle\pm1.28}$ & $-0.15$ & $+4.79$ \\
Texture & $11.47{\scriptstyle\pm2.75}$ & $12.21{\scriptstyle\pm2.21}$ & $+0.74$ & $+2.28$ & $11.01{\scriptstyle\pm1.84}$ & $-0.46$ & $+2.86$ & $11.75{\scriptstyle\pm1.69}$ & $+0.28$ & $+3.76$ \\
Openimage-O & $11.55{\scriptstyle\pm2.08}$ & $12.27{\scriptstyle\pm1.85}$ & $+0.73$ & $+2.02$ & $11.16{\scriptstyle\pm1.68}$ & $-0.38$ & $+1.84$ & $11.83{\scriptstyle\pm1.51}$ & $+0.28$ & $+3.12$ \\
\rowcolor{gray!20}\textit{Average} & $12.31{\scriptstyle\pm2.48}$ & $13.00{\scriptstyle\pm2.20}$ & $+0.69$ & $+2.17$ & $11.85{\scriptstyle\pm1.96}$ & $-0.47$ & $+2.01$ & $12.55{\scriptstyle\pm1.87}$ & $+0.24$ & $+2.86$ \\
\bottomrule
\end{tabular}
}
\label{tab:ablation}
\end{table}
}

\newcommand{\figresiduals}{%
\begin{figure}[t!]
    \centering
    \begin{subfigure}{\linewidth}
        \centering
        \includegraphics[width=\linewidth]{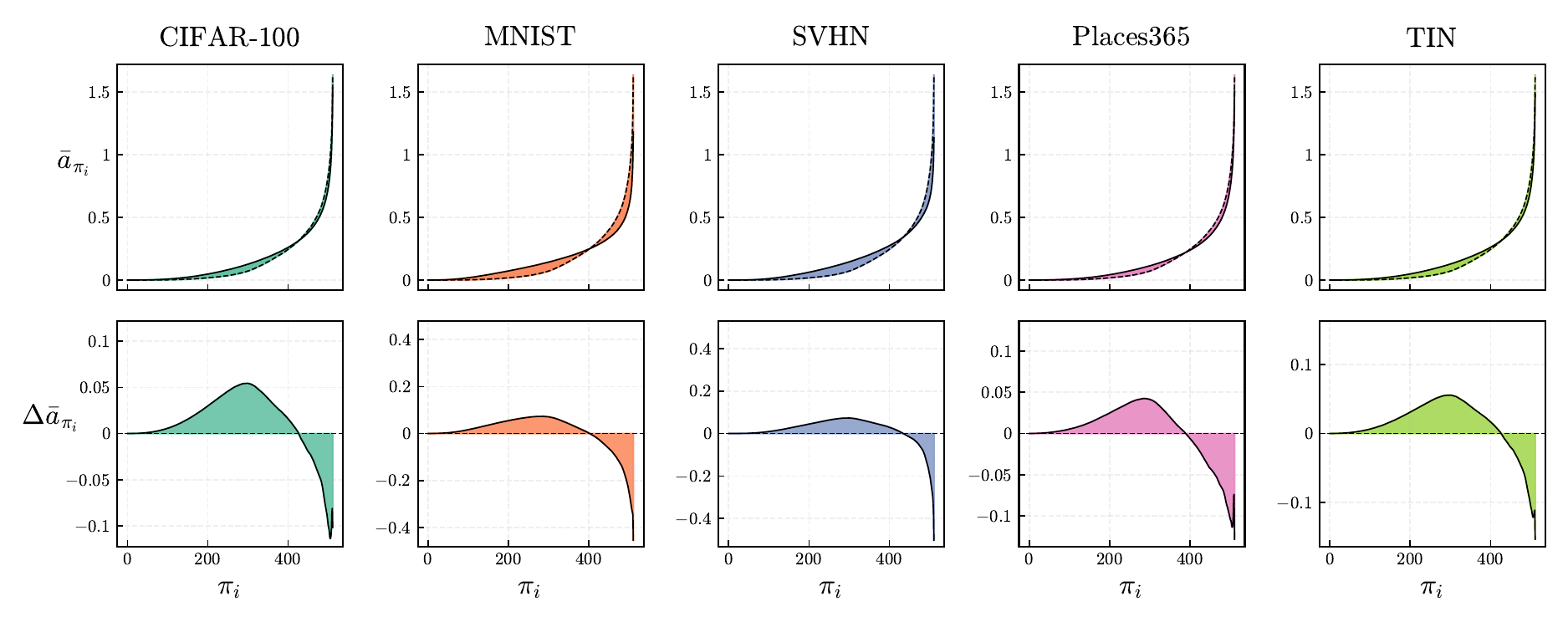}
        \caption{ID dataset: CIFAR-10. Model: ResNet-18.}
        \label{subfig:combined_a}
    \end{subfigure}

    \begin{subfigure}{\linewidth}
        \centering
        \includegraphics[width=\linewidth]{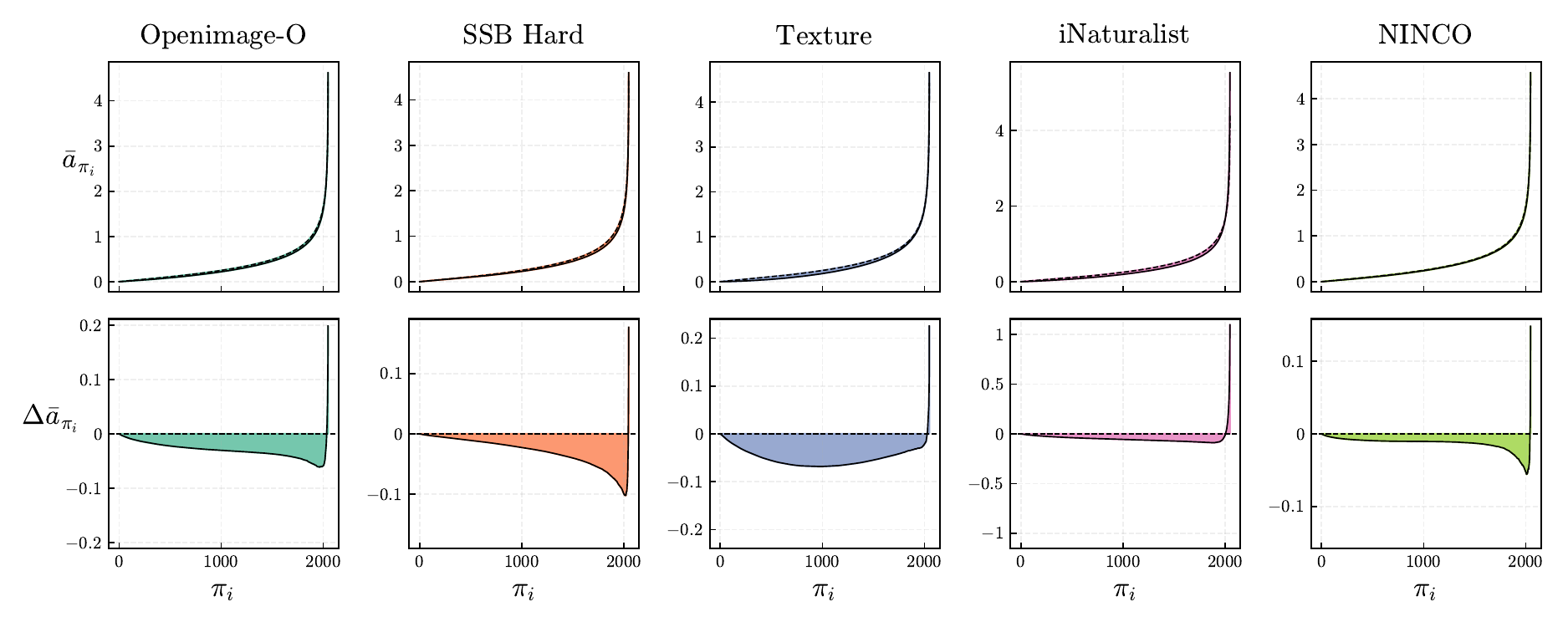}
        \caption{ID dataset: ImageNet. Model: ResNet-50.}
        \label{subfig:combined_b}
    \end{subfigure}
    \caption{Comparison of activation profiles for ID (dashed line) and OoD datasets (solid line). $a_{\pi_i}$ shows the average ranked activations, while $\Delta a_{\pi_i}$ indicates the residual difference between the sorted ID and OoD activations. Two distinct behaviors emerge across benchmarks: in the CIFAR-10 case, OoD samples consistently exhibit weaker high-rank activations than ID; in the ImageNet case, the opposite holds across all OoD datasets.}
    \label{fig:residuals}
\end{figure}
}

\usepackage[accsupp]{axessibility}  

\usepackage[pagebackref,breaklinks,colorlinks,citecolor=eccvblue]{hyperref}
\usepackage{hyperref}

\usepackage{orcidlink}

\begin{document}

\title{Ranked Activation Shift for Post-Hoc Out-of-Distribution Detection} 

\titlerunning{Ranked Activation Shift}

\author{Gianluca Guglielmo \and Marc Masana}

\authorrunning{G.~Guglielmo and M.~Masana}


\institute{{Graz University of Technology}\\
{\tt\small \{guglielmo, mmasana\}@tugraz.at}
}
\maketitle

\begin{abstract}
State-of-the-art post-hoc out-of-distribution detection methods rely on intermediate layer activation editing. However, they exhibit inconsistent performance across datasets and models. We show that this instability is driven by differences in the activation distributions, and identify a failure mode of scaling-based methods that arises when penultimate layer activations are not rectified. Motivated by this analysis, we propose \ours, a hyperparameter-free post-hoc method that replaces sorted activation magnitudes with a fixed in-distribution reference profile. Our simple plug-and-play method shows strong and consistent performance across datasets and architectures without assumptions on the penultimate layer activation function, and without requiring any hyperparameter tuning, while empirically preserving in-distribution classification accuracy. We further analyze what drives the improvement, showing that both inhibiting and exciting activation shifts independently contribute to better out-of-distribution discrimination\footnote{Code is available at:  \url{https://github.com/gigug/RAS}.}.
\end{abstract}

\section{Introduction}
\label{sec:introduction}
Models deployed in real-world scenarios often face data that differs from what they were trained on. As a result, they must be able to identify such inputs and respond appropriately~\cite{amodei2016concrete, yang2024generalized, miyaigeneralized}. This problem, known as out-of-distribution~(OoD) detection, is essential for building safe and reliable AI systems. Without these safety mechanisms, model predictions cannot be fully trusted, especially in high-stakes applications such as autonomous driving, medical imaging, or financial decision-making, where overconfident errors can have serious consequences~\cite{hong2024out, yang2024generalized, hendrycks2022scaling, thimonier2024comparative}.

\tabcompare{}

In recent years, several OoD detection methods have been proposed, covering a variety of methodological families~\cite{yang2024generalized, miyaigeneralized}. This paper focuses on \textit{post-hoc} OoD detection methods, which operate without the need for retraining past the initial classification task. This provides a strong advantage since the models can be directly integrated into existing pipelines with very low computational costs. Established post-hoc methods typically exploit either the penultimate layer activations or the model logits, which naturally exhibit useful information to detect OoD samples~\cite{hendrycks2018baseline, liu2020energy, yang2024generalized}.

\textit{Score-enhancing} post-hoc OoD detection methods improve the discriminative power of other logit-based OoD detection methods~\cite{liu2023gen}. They establish a rule to adjust intermediate layer activations to suppress noise hindering OoD detection and amplify relevant signals. Among these, we find ReAct~\cite{sun2021react}, DICE~\cite{sun2022dice}, ASH~\cite{djurisic2023extremely} and SCALE~\cite{xu2024scaling}. Although these methods perform well on a variety of settings, their performance tends to vary depending on the specific model architecture and datasets. Moreover, they require hyperparameter optimization, typically via access to an outlier dataset. This adds an additional layer of complexity to the task and ties the performance of the method to the correct choice of hyperparameters. We present a comparison of scoring and score-enhancing methods in~\cref{tab:compare}.

In this paper, we present a \textbf{hyperparameter-free, score-enhancing OoD detection method} which performs robustly and consistently on different datasets and architectures by shifting the penultimate layer activations to match the mean ranked activation intensities found in ID data. 

Our contributions can be summarized as follows:
\begin{itemize}
    \item we show where scaling-based enhancing methods underperform, including a failure mode arising from unrectified activations,
    \item we introduce \ours, a novel universal hyperparameter-free enhancing method that replaces ranked activations with a fixed ID reference profile, and
    \item we analyze what drives \ours's improvement, showing that, contrary to the prevailing assumption, both inhibitory and excitatory activation shifts independently contribute to OoD separation.
\end{itemize}

\section{Preliminaries}
\label{Preliminaries}
\subsubsection{Terminology.}
We consider a standard supervised classification setting with \mbox{$\mathbf{x}_i \in \mathcal{X}$} representing the input space, and \mbox{$y_i \in \mathcal{Y} = \{1, \dots, C\}$, $\mathcal{Y}$} representing the labels of $C$ known classes. Model $f_\theta$ is trained on an ID dataset $\mathcal{D}_{\text{in}} = \{(\mathbf{x}_i, y_i)\}_{i=1}^n$, which is drawn from an unknown distribution $\mathcal{P_{\text{in}}} \in \mathcal{X} \times \mathcal{Y}$. During inference, the model processes samples drawn from a mixture of $\mathcal{P}_{\text{in}}$ and an external $\mathcal{P}_{\text{out}}$. The goal of OoD detection is to define a scoring function $S(\mathbf{x})$ that maps a given input sample to a scalar value representing the confidence that $\mathbf{x}$ is drawn from $\mathcal{P}_{\text{in}}$. As introduced by~\cite{hendrycks2018baseline}, a well constructed $S(\mathbf{x})$ should assign higher values to ID samples and lower values to OoD samples, such that a threshold $\tau$ can be chosen to effectively separate the two distributions.
Consequently, the binary detector $g(\mathbf{x})$ is defined as:
\begin{equation}
    g(\mathbf{x}) = \begin{cases} 
0 & \text{if } S(\mathbf{x}) \ge \tau \\
1 & \text{if } S(\mathbf{x}) < \tau\,,
\end{cases}
\end{equation}
where $0$ indicates a sample being OoD, while $1$ indicates ID. The threshold $\tau$ can be adjusted depending on the desired balance between sensitivity and specificity for OoD detection.

\subsubsection{Post-hoc methods} operate without the need for retraining past the classification task. Generally this allows for strong OoD detection capabilities without a trade-off with ID classification accuracy~\cite{borlino2024foundation}. The de-facto OoD detection baseline, \textit{Maximum Softmax Probability} (MSP)~\cite{hendrycks2018baseline}, relies on the maximum softmax output as the OoD score. EBO~\cite{liu2020energy} improves this by defining an energy function that takes the whole logit vector as input. GEN~\cite{liu2023gen} further pushes this by reweighting the softmax distribution, aggregating only the top predicted classes to produce a more discriminative score. Distance-based methods~\cite{venkataramanan2023gaussian, sun2022out} quantify how far a sample’s intermediate-layer representation deviates from the ID representations, flagging it as OoD if the distance reaches a certain threshold. Some methods combine intermediate representations and logits information through a single unified score~\cite{yang2024generalized}. ViM~\cite{haoqi2022vim} does it by defining an additional \textit{virtual logit} equal to the sample distance from the ID-based principal subspace.

\subsubsection{Score-enhancing methods} act on intermediate activations to enhance logit-based methods and facilitate OoD detection at inference time. DICE~\cite{sun2022dice} introduces a directed sparsification strategy for OoD detection, pruning weights in the penultimate layer based on their contribution to the output. ReAct~\cite{sun2021react} is based on the empirical observation that OoD samples often trigger abnormally high activation values in the penultimate layer, leading to overconfident predictions. ReAct truncates these activations using a fixed threshold $c$, computed as a percentile of the activations from the ID training data. Conversely, ASH~\cite{djurisic2023extremely} simplifies the representation by removing a large portion of the penultimate layer activations. The authors introduce three versions: simply prune (ASH-P), prune and binarize (ASH-B) or prune and scale (ASH-S). SCALE~\cite{xu2024scaling} builds a theoretical framework to show why ASH~\cite{djurisic2023extremely} is effective and introduces a variation of it that scales activations without the need for pruning.

\section{To prune or to scale?}
\label{sec:motivation}
ASH~\cite{djurisic2023extremely} and SCALE~\cite{xu2024scaling} rely heavily on auxiliary OoD data and achieve strong performance across many benchmarks~\cite{zhang2024openood}. Nonetheless, they have some unaddressed limitations, which we explore in this section. Specifically, we start by examining whether the choice between scaling and pruning is as straightforward as assumed, and we conclude by identifying a failure mode of scaling that occurs when the penultimate layer activations are not rectified.

Given an activation\footnote{We refer to the output of a layer as activations or feature vector, interchangeably.} vector $\mathbf{a}$, the following quantities are defined as in~\cite{xu2024scaling}: 
\begin{equation}
    r = \frac{Q}{Q_p}, \quad \text{where } Q = \sum_{j=1}^D a_j \textrm{ and } Q_p = \sum_{a_j > P_p(\mathbf{a})}^D a_j\,,
\end{equation}
\noindent where $P_p(\mathbf{a})$ is the $p$-th percentile of $\mathbf{a}$. SCALE applies a scaling factor $e^r$ to the whole vector $\mathbf{a}$, which is then passed to the final classifier:
\begin{equation}
    \mathbf{z} = \mathbf{W}(e^r \mathbf{a}) + \mathbf{b}\,.
\end{equation}

ASH-S~\cite{djurisic2023extremely} only scales the activations that are greater than $P_p(\mathbf{a})$ by $e^r$ , pruning the remainder to zero. The rationale behind SCALE's choice stems from the following two assumptions: a) the mean of ID activations is higher than that of OoD activations, with the variances remaining equivalent, and b) the penultimate layer activations can be modeled by rectified Gaussian distributions (an assumption also found in the theoretical analysis of ReAct~\cite{sun2021react}).

With these assumptions, by defining ID activations as $\mathbf{a}_j^{(\text{ID})} \sim \mathcal{N}^R(\mu^{\text{ID}}, \sigma^{\text{ID}})$ and OoD ones as $\mathbf{a}_j^{(\text{OoD})} \sim \mathcal{N}^R(\mu^{\text{OoD}}, \sigma^{\text{OoD}})$, SCALE proves that if condition
\begin{equation}
\label{condition_mu}
    \frac{\mu^{\text{ID}}}{\sigma^{\text{ID}}}>\frac{\mu^{\text{OoD}}}{\sigma^{\text{OoD}}}
\end{equation} 
is met, then there exists a percentile $p$ such that:
\begin{equation}
\label{condition_q}
    \frac{Q_p^{\text{ID}}}{Q^{\text{ID}}} < \frac{Q_p^{\text{OoD}}}{Q^{\text{OoD}}}\,.
\end{equation}
In turn, they prove that Condition \ref{condition_q} implies that \textit{pruning} activations is detrimental to OoD detection, while \textit{scaling} effectively helps it; when the condition is reversed, so are these effects too.

We empirically examine whether Condition~\ref{condition_mu} holds across standard benchmarks by defining the following quantity and studying its sign:
\begin{equation}
    \Gamma(p; \mathcal{{D}}_{{\text{{in}}}}, \mathcal{{D}}_{{\text{{out}}}}) = \frac{Q_p^{\text{OoD}}}{Q^{\text{OoD}}} - \frac{Q_p^{\text{ID}}}{Q^{\text{ID}}}\,.
\end{equation}

\begin{figure}[b!]
    \begin{subfigure}{\linewidth}
    \centering
    \includegraphics[width=\linewidth]{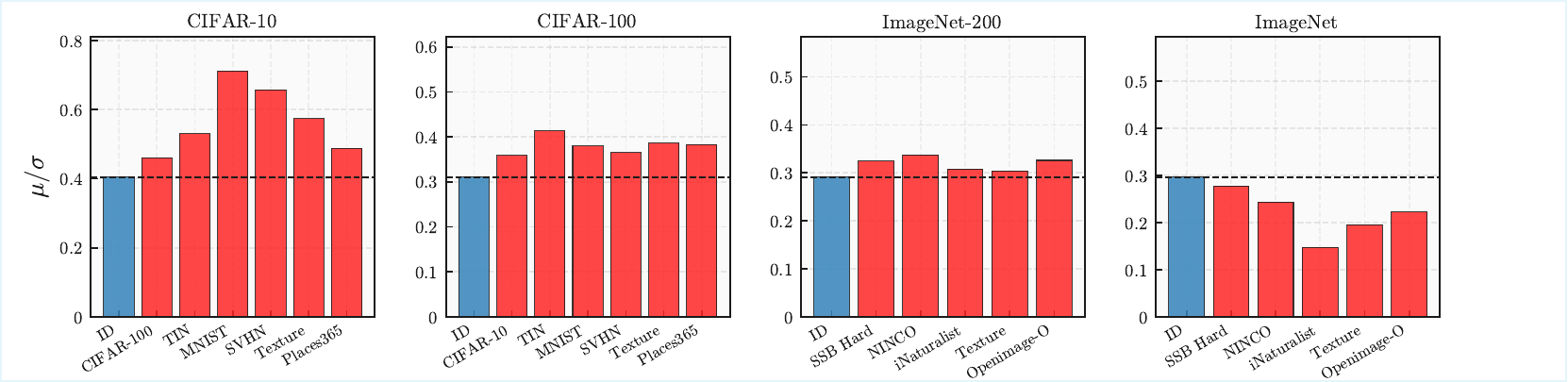}
    \caption{Pre-ReLU activations; ID is shown in blue; OoD are shown in red.}
    \label{subfig:a}
    \end{subfigure}
    \begin{subfigure}{\linewidth}
    \centering
    \includegraphics[width=\linewidth]{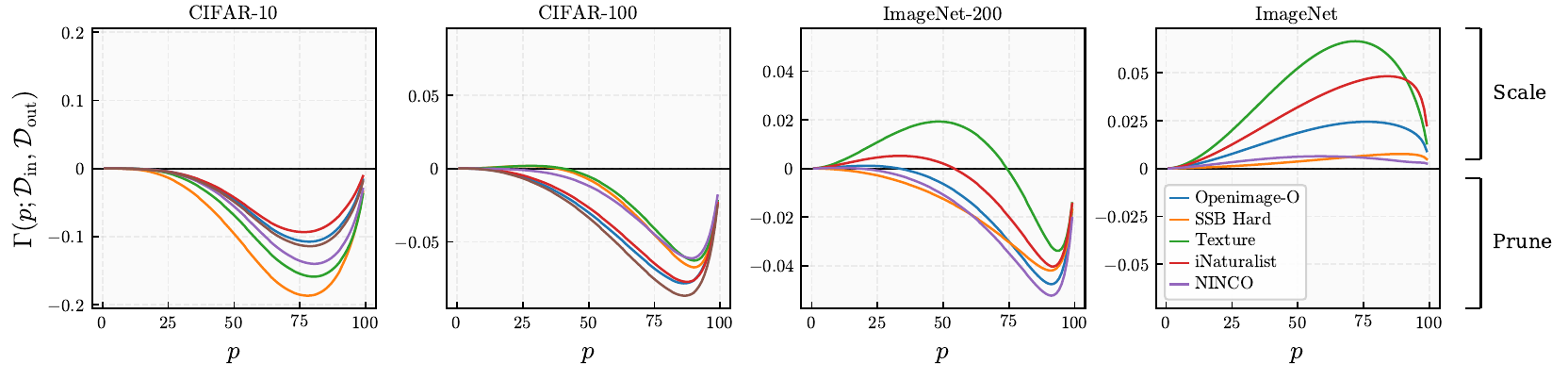}
    \caption{$\Gamma(p; \mathcal{{D}}_{{\text{{in}}}}, \mathcal{{D}}_{{\text{{out}}}})$ for percentiles $p$. OoD datasets differentiated by colors.}
    \label{subfig:b}
    \end{subfigure}
    \caption{The condition required by SCALE's assumptions to hold is violated in three of the four primary OpenOOD setups. In these setups $\mu_{ID}/\sigma_{ID} < \mu_{OoD}/\sigma_{OoD}$, which leads to pruning being more discriminative than scaling for certain percentiles.}
    \label{fig:myfig}
\end{figure}

The four primary setups from OpenOOD~\cite{zhang2024openood} all employ CNN-based architectures with rectification at the penultimate layer (details in \cref{sec:experiments}). We compute $\mu/\sigma$ for the ID and OoD datasets of each setup and show in \cref{subfig:a} that Condition~\ref{condition_mu} is not met for three of the four setups. 

We empirically evaluate the implication~\ref{condition_mu}$\implies$\ref{condition_q} in \cref{subfig:b}, which plots $\Gamma(p; \mathcal{{D}}_{{\text{{in}}}}, \mathcal{{D}}_{{\text{{out}}}})$ across percentiles $p$ for the same datasets: when Condition~\ref{condition_mu} is false, $\Gamma(p; \mathcal{{D}}_{{\text{{in}}}}, \mathcal{{D}}_{{\text{{out}}}})$ is negative for most values of $p$. Consequently, for three of the four benchmarks, pruning activations is more discriminative than scaling them for a broad range of percentiles. The further this quantity deviates from zero, the stronger the advantage of either pruning or scaling, while for percentiles closer to zero the difference between the two becomes negligible. Since $p$ is a hyperparameter, its selection is critical: the relative efficacy of pruning and scaling is highly sensitive to its choice.

\subsection{Absence of rectification}
As shown in \cref{table:model_relu}, many modern architectures do not apply any rectification to their penultimate layer activations, including both convolutional models such as ConvNeXt~\cite{liu2022convnet} and transformer-based models such as ViT~\cite{dosovitskiy2020vit}. This has direct consequences for the scaling factor $r$, used in SCALE and ASH-S. When activations are rectified, $Q_p \leq Q$ guarantees $r \geq 1$. However, in the absence of rectification, activations can take negative values, meaning $Q$ may be small or even negative, and $r$ is no longer constrained to be positive or greater than one. This breaks the monotonic relationship between $r$ and the pruning threshold $p$ that the theoretical analysis relies upon, and the interpretation of scaling as a strictly amplifying operation no longer holds.

\begin{table}[t!]
\centering
\caption{Best performing ImageNet models~\cite{Bekhouche2025ImageNetLeaderboard} rarely have activations be all positive.}
\label{table:model_relu}
\setlength{\tabcolsep}{8pt}
\begin{tabular}{lcccr}
\toprule
\textbf{Model} & \textbf{Top-1 Acc.} & \textbf{Activation} & $a_{i}\geq0,\ \forall i$ \\
\midrule
ConvNeXt-XL~\cite{liu2022convnet} & 87.76\% & GELU & \xmark \\
DeiT-S~\cite{deit2020}            & 87.76\% & GELU & \xmark \\
ConvNeXt-L~\cite{liu2022convnet}  & 87.46\% & GELU & \xmark \\
ConvNeXt-B~\cite{liu2022convnet}  & 86.82\% & GELU & \xmark \\
DeiT-B~\cite{deit2020}            & 85.67\% & GELU & \xmark \\
EfficientNet-B6~\cite{tan2019efficientnet} & 83.93\% & SiLU & \xmark \\
RegNet-Y-640~\cite{radosavovic2020designing} & 83.83\% & ReLU & \cmark \\
EfficientNet-B5~\cite{tan2019efficientnet} & 83.36\% & SiLU & \xmark \\
CvT-13~\cite{wu2021cvt}           & 82.79\% & GELU & \xmark \\
ResNet-152~\cite{he2016deep}      & 82.50\% & ReLU & \cmark \\
\bottomrule
\end{tabular}
\end{table}

\section{Ranked Activation Shift}
\label{sec:method}
\begin{figure}[t]
    \centering
    \includegraphics[width=\linewidth]{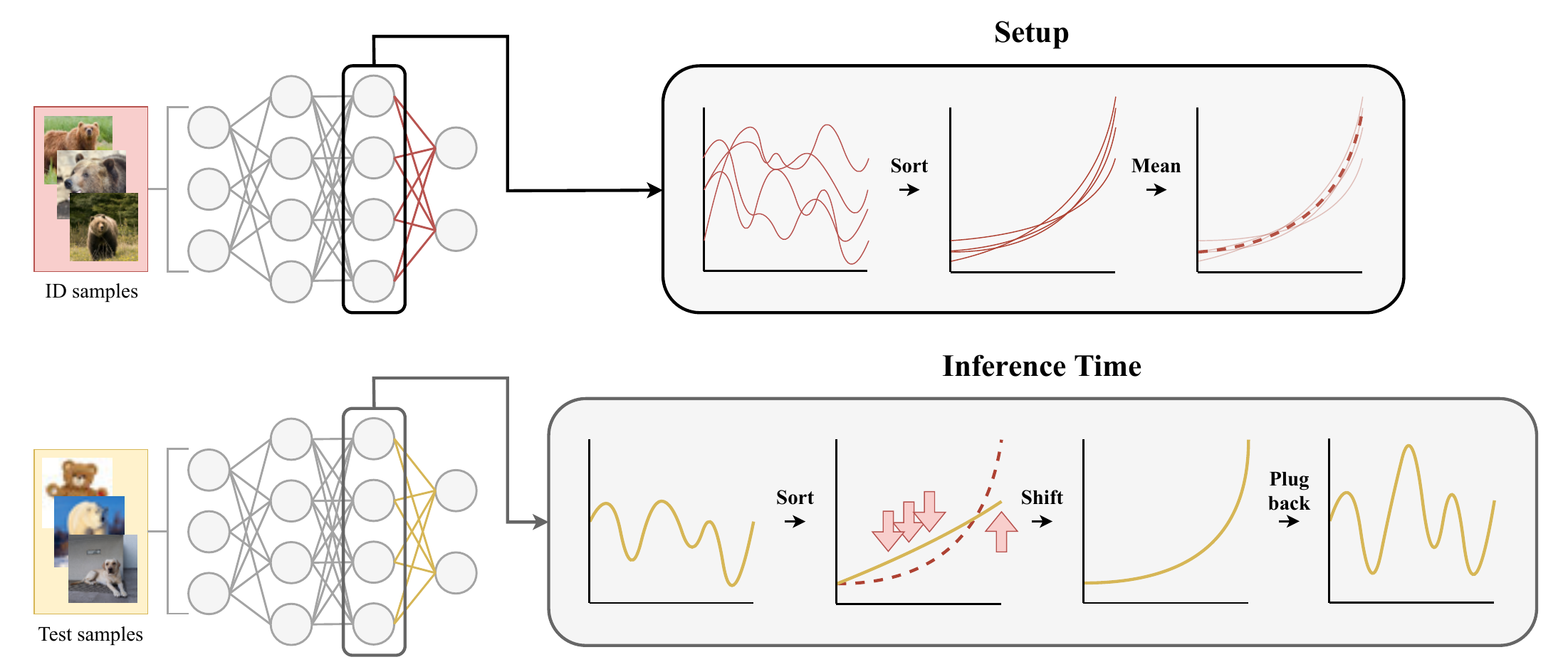}
    \caption{Overview of \ours. At \textit{setup} time, \ours computes a reference vector $\boldsymbol{\mu}$ by sorting each ID train sample by intensity and then averaging across all. At \textit{inference} time, each sample's activations are shifted so that their ranked activation intensities match $\boldsymbol{\mu}$. Subsequently, they are passed to the classifier and the chosen OoD scorer is used.}
    \label{fig:method_depiction}
\end{figure}

Current post-hoc methods~\cite{sun2021react, sun2022dice, djurisic2023extremely, xu2024scaling} rely on threshold-based clipping or pruning, which does not leverage the global shape of the activation vector. We propose Ranked Activation Shift~(\ours), which compares the sorted activations of new samples against an ID reference, as depicted in \cref{fig:method_depiction}. During a one-time offline phase, we compute a reference profile vector $\boldsymbol{\mu} \in \mathbb{R}^d$, defined as the mean of the ranked\footnote{We refer to the terms \textit{ranking} and \textit{sorting}, interchangeably.} activation vectors over a subset of ID data:
\begin{equation}
\boldsymbol{\mu} = \frac{1}{N} \sum_{i=1}^N \mathrm{r}(\mathbf{a}_i),
\end{equation}
where $a_i$ are the penultimate layer activations extracted from samples $\mathbf{x}_i \in \mathcal{\mathcal{D}_{\text{in}}}$, while $\mathrm{r}(\cdot)$ is a function that arranges the elements of the vector in ascending order. As shown in~\cref{fig:subsampling}, the reference profile can be successfully built using few samples, with the option of exploiting an held-out in-distribution dataset that was not seen during training.

At inference time, for a given input $\mathbf{x} \in \mathcal{\mathcal{D}_{\text{in}}} \cup \mathcal{\mathcal{D}_{\text{out}}}$, we extract its penultimate layer activations $\mathbf{a}$. Let $\pi$ be the permutation of indices $\{1, \dots, d\}$ that sorts $\mathbf{a}$ in ascending order, such that $s_j = a_{\pi(j)}$ corresponds to the $j$-th largest activation. Next, we construct the modified activation vector $\bar{\mathbf{a}}$ by assigning the reference value $\mathbf{\mu}_j$ to the original position $\pi(j)$:
\begin{equation}
\bar{a}_{\pi(j)} = \mu_j \quad \text{for } j=1, \dots, d\,.
\end{equation}

\begin{algorithm}[b!]
\caption{\ours: Ranked Activation Shift}
\label{alg:wt}
\begin{minipage}{0.75\linewidth}
\begin{algorithmic}[1]
\Require ID training set $\mathcal{D}_{\text{in}}$, feature dimension $d$
\Ensure Modified activation vector $\bar{\mathbf{a}}$ at inference-time

\noindent\textbf{Setup}
\State Initialize reference vector $\boldsymbol{\mu} \leftarrow \mathbf{0} \in \mathbb{R}^d$
\For{each $\mathbf{x}_i \in \mathcal{D}_{\text{in}}$}
    \State Extract activation vector $\mathbf{a}_i$
    \State $\boldsymbol{\mu} \leftarrow \boldsymbol{\mu} + \mathrm{r}(\mathbf{a}_i)$
\EndFor
\State $\boldsymbol{\mu} \leftarrow \boldsymbol{\mu} / |\mathcal{D}_{\text{in}}|$

\noindent\textbf{Inference-Time Substitution}
\State Extract activation vector\footnote{In ViTs, $\mathbf{a}$ is the output representation of the class token.} $\mathbf{a} \in \mathbb{R}^d$ for input $\mathbf{x}$
\State Compute permutation $\pi$ such that $a_{\pi(1)} \ge \cdots \ge a_{\pi(d)}$
\State Initialize $\bar{\mathbf{a}} \leftarrow \mathbf{a}$
\For{$j = 1$ to $d$}
\State $\bar{a}_{\pi(j)} \leftarrow \mu_j$
\EndFor
\State Continue feed-forward, compute chosen $S(\mathbf{x})$
\end{algorithmic}
\end{minipage}
\end{algorithm}

By strictly imposing the value distribution of $\boldsymbol{\mu}$ onto the input, we effectively perform a histogram matching step. This ensures that $\bar{\mathbf{a}}$ exhibits the average statistical profile of ID data, while preserving the spatial orientation of the original input $\mathbf{x}$. Consequently, if an OoD sample produces an anomalous activation distribution (e.g., peaked noise or heavy tails), this substitution forces it into an ID-like range before classification. 
The step-by-step process is shown in Alg.~\ref{alg:wt}.

\section{Experiments}
\label{sec:experiments}
We evaluate \ours using the OpenOOD benchmark~\cite{zhang2024openood}, a standardized framework to compare OoD detection methods. It provides a unified pipeline with consistent data splits and pre-trained model backbones, ensuring a fair comparison between methods. We provide benchmark results on all available ID datasets and their corresponding models in OpenOOD: ImageNet~\cite{deng2009imagenet} with ResNet50~\cite{he2016deep}, \mbox{CIFAR-10}, \mbox{CIFAR}-100~\cite{krizhevsky2009learning} and ImageNet-200 (a subset of ImageNet) with ResNet18~\cite{he2016deep}. We further extend our evaluation to include EfficientNet~\cite{tan2019efficientnet}, which does not use residual blocks, ConvNeXt~\cite{liu2022convnet}, which aggregates different strategies to modernize convolutional neural networks, and vision-transformers, namely ViT-B/16~\cite{dosovitskiy2020vit} and Swin-T~\cite{liu2021swin}. For these models not available in the OpenOOD framework, we use checkpoints from Torchvision~\cite{torchvision2016} and HuggingFace~\cite{wolf2020transformers}.

The evaluation protocol follows the splits summarized in \cref{tab:datasets}. For each experiment, ID is represented by the corresponding test set of the pretrained backbone's dataset, containing unseen samples drawn from the same distribution. OoD datasets are divided into near- and far-OoD subsets, which indicates the semantic distance from the ID dataset~\cite{yang2024generalized, zhang2024openood}.

\begin{table}[t]
    \centering
    \caption{OoD evaluation splits for each ID dataset.\label{tab:datasets}}
    \resizebox{\linewidth}{!}{%
    \renewcommand{\arraystretch}{1.2}
    \begin{tabular}{c@{\hskip 0.2in}c@{\hskip 0.2in}c@{\hskip 0.2in}c@{\hskip 0.2in}c}
    \toprule
    & \textbf{CIFAR-10}~\cite{krizhevsky2009learning}
    & \textbf{CIFAR-100}~\cite{krizhevsky2009learning}
    & \textbf{ImageNet-200}~\cite{deng2009imagenet}
    & \textbf{ImageNet}~\cite{deng2009imagenet} \\
    \midrule
    \multirow{2}{*}{\textbf{Near-OoD}}
    & CIFAR-100~\cite{krizhevsky2009learning}       & CIFAR-10~\cite{krizhevsky2009learning}       & SSB-Hard~\cite{zhang2024openood}   & SSB-Hard~\cite{zhang2024openood} \\
    & TinyImageNet~\cite{deng2009imagenet} & TinyImageNet~\cite{deng2009imagenet} & NINCO~\cite{bitterwolf2023or} & NINCO~\cite{bitterwolf2023or} \\
    \midrule
    \multirow{4}{*}{\textbf{Far-OoD}}
    & Texture~\cite{cimpoi2014describing}  & Texture~\cite{cimpoi2014describing}    & Texture~\cite{cimpoi2014describing}                                    & Texture~\cite{cimpoi2014describing} \\
    & MNIST~\cite{deng2012mnist}         & MNIST~\cite{deng2012mnist}      & iNaturalist~\cite{van2018inaturalist} & iNaturalist~\cite{van2018inaturalist} \\
    & SVHN~\cite{netzer2011reading}      & SVHN~\cite{netzer2011reading}       & OpenImage-O~\cite{haoqi2022vim}           & OpenImage-O~\cite{haoqi2022vim} \\
    & Places365~\cite{zhou2017places}  & Places365~\cite{zhou2017places} &                                                  &  \\
    \bottomrule
    \end{tabular}}
\end{table}

We employ the most commonly used OoD detection metrics~\cite{hendrycks2018baseline, miyaigeneralized, borlino2024foundation, yang2024generalized} for these scenarios: AUROC, the area under the receiver operating characteristic curve (the higher the better); AUPR, the area under the precision-recall curve (the higher the better); and FPR@TPR95, the false-positive rate when the true-positive rate is 95\% (the lower the better).

\subsection{Main benchmark results}
We start by comparing RAS with state-of-the-art enhancing post-hoc OoD methods. We follow their original implementations by applying each method to the penultimate layer and using the EBO function to compute the final score $S(\mathbf{x})$. In the case of vision-transformers, \ours is applied to the class token, since the other penultimate layer tokens are discarded during classification.

\Cref{tab:full_comparison_table} summarizes the results for various ID datasets and architectures. \ours consistently outperforms every compared methods on all but one scenario (ImageNet with ResNet50), where it still achieves a strong improvement over the standard EBO performance. As expected, ASH and SCALE fail on architectures whose penultimate layer contains negative-valued activations (ConvNeXt, ViT-B/16). It is crucial to note that \ours achieves these results without requiring any hyperparameter optimization. In contrast, the other enhancing methods require a hyperparameter (ID-OoD threshold) tuning on a held-out OoD dataset, with the reported values representing their optimal configurations. This underlines the robustness of \ours, as it improves over or remains competitive against methods displaying their peak potential performance. Extended results in Supplementary.

\begin{table}[t]
\centering
\caption{Across a comprehensive range of dataset--architecture combinations, \ours (hyperparameter-free) consistently exhibits best or competitive performance compared to state-of-the-art enhancing post-hoc OoD detection techniques on their optimized hyperparameter values. \textbf{Bold} indicates best result, \underline{underline} second and third.}
\resizebox{\linewidth}{!}{
\begin{tabular}{l cc cc cc cc cc cc cc cc}
\toprule
& \multicolumn{2}{c}{\textbf{CIFAR-10}} & \multicolumn{2}{c}{\textbf{CIFAR-100}} & \multicolumn{2}{c}{\textbf{IN-200}} & \multicolumn{10}{c}{\textbf{ImageNet}} \\
\cmidrule(lr){2-3} \cmidrule(lr){4-5} \cmidrule(lr){6-7} \cmidrule(lr){8-17}

& \multicolumn{2}{c}{\textbf{ResNet18}} & \multicolumn{2}{c}{\textbf{ResNet18}} & \multicolumn{2}{c}{\textbf{ResNet18}} & \multicolumn{2}{c}{\textbf{ResNet50}} & \multicolumn{2}{c}{\textbf{EffNet-B0}} & \multicolumn{2}{c}{\textbf{ConvNx-T}} & \multicolumn{2}{c}{\textbf{ConvNx-B}} & \multicolumn{2}{c}{\textbf{ViT-B/16}} \\

\cmidrule(lr){2-3} \cmidrule(lr){4-5} \cmidrule(lr){6-7} \cmidrule(lr){8-9} \cmidrule(lr){10-11} \cmidrule(lr){12-13} \cmidrule(lr){14-15} \cmidrule(lr){16-17} & AUC$\uparrow$ & FPR$\downarrow$ & AUC$\uparrow$ & FPR$\downarrow$ & AUC$\uparrow$ & FPR$\downarrow$ & AUC$\uparrow$ & FPR$\downarrow$ & AUC$\uparrow$ & FPR$\downarrow$ & AUC$\uparrow$ & FPR$\downarrow$ & AUC$\uparrow$ & FPR$\downarrow$ & AUC$\uparrow$ & FPR$\downarrow$ \\
\midrule
EBO & $\underline{90.00}$ & $\underline{48.24}$ & $80.15$ & $56.26$ & $87.51$ & $45.01$ & $84.03$ & $50.46$ & $\underline{76.76}$ & $\underline{72.29}$ & $\underline{69.36}$ & $\underline{83.03}$ & $59.36$ & $93.29$ & $72.35$ & $88.48$ \\
\midrule
ReAct & $89.31$ & $\underline{51.12}$ & $80.52$ & $54.93$ & $88.13$ & $42.10$ & $87.15$ & $42.46$ & $\mathbf{84.27}$ & $\underline{49.01}$ & $\underline{77.87}$ & $\underline{69.45}$ & $\underline{80.85}$ & $\underline{64.06}$ & $\underline{79.12}$ & $\underline{66.15}$ \\
DICE & $82.27$ & $57.85$ & $79.80$ & $56.82$ & $87.19$ & $46.66$ & $83.80$ & $54.07$ & $43.32$ & $91.33$ & $57.78$ & $93.86$ & $58.10$ & $95.30$ & $\underline{75.53}$ & $\underline{65.23}$ \\
ASH-P & $\underline{89.47}$ & $52.55$ & $80.34$ & $55.77$ & $88.02$ & $43.44$ & $85.44$ & $46.04$ & $63.24$ & $80.79$ & $31.68$ & $99.26$ & $16.89$ & $99.80$ & $22.32$ & $98.74$ \\
ASH-B & $77.50$ & $81.48$ & $79.79$ & $61.29$ & $89.24$ & $42.48$ & $\underline{88.75}$ & $\underline{37.11}$ & $47.50$ & $86.81$ & $31.01$ & $98.05$ & $\underline{65.33}$ & $79.84$ & $52.57$ & $95.51$ \\
ASH-S & $82.90$ & $77.47$ & $\underline{81.52}$ & $\underline{54.29}$ & $\mathbf{90.51}$ & $\underline{38.15}$ & $\underline{89.68}$ & $\underline{35.05}$ & $56.18$ & $81.21$ & $32.59$ & $97.64$ & $26.86$ & $99.67$ & $22.68$ & $98.71$ \\
SCALE & $85.18$ & $71.77$ & $\underline{81.29}$ & $\underline{54.60}$ & $\underline{90.29}$ & $\underline{38.89}$ & $\mathbf{90.42}$ & $\mathbf{34.02}$ & $59.05$ & $79.70$ & $60.10$ & $84.32$ & $65.22$ & $\underline{76.53}$ & $68.64$ & $89.52$ \\
\rowcolor{gray!20}\textbf{\ours} & $\mathbf{90.24}$ & $\mathbf{40.16}$ & $\mathbf{82.09}$ & $\mathbf{52.31}$ & $\underline{89.51}$ & $\mathbf{36.70}$ & $86.55$ & $40.92$ & $\underline{83.78}$ & $\mathbf{46.87}$ & $\mathbf{83.31}$ & $\mathbf{46.28}$ & $\mathbf{84.86}$ & $\mathbf{45.29}$ & $\mathbf{81.48}$ & $\mathbf{55.19}$ \\
\bottomrule
\end{tabular}
}
\label{tab:full_comparison_table}
\end{table}

\subsubsection{Preserving ID accuracy} is critical for OoD detection methods, especially for enhancing post-hoc ones, as these can directly interfere with the forward pass. As shown in \cref{tab:tab_acc}, \ours keeps ID accuracy virtually unchanged.

\begin{table}[h]
\centering
\caption{RAS empirically preserves ID accuracy.}
\label{tab:tab_acc}
\begin{tabular}{l c c c c c c c c}
\toprule
& \textbf{C-10} & \textbf{C-100} & \textbf{IN-200} & \multicolumn{5}{c}{\textbf{ImageNet}} \\
\cmidrule(lr){2-2} \cmidrule(lr){3-3} \cmidrule(lr){4-4} \cmidrule(lr){5-9}

& \multicolumn{1}{c}{\textbf{RN18}} & \multicolumn{1}{c}{\textbf{RN18}} & \multicolumn{1}{c}{\textbf{RN18}} & \multicolumn{1}{c}{\textbf{RN50}} & \multicolumn{1}{c}{\textbf{ENB0}} & \multicolumn{1}{c}{\textbf{CNx-T}} & \multicolumn{1}{c}{\textbf{CNx-B}} & \multicolumn{1}{c}{\textbf{ViT-B/16}} \\
\midrule
Orig. & $95.06$ & $77.25$ & $86.37$ & $76.17$ & $77.80$ & $82.11$ & $83.74$ & $81.14$  \\
RAS & $95.03$ & $77.26$ & $86.16$ & $76.04$ & $77.73$ & $82.15$ & $83.71$ & $81.14$ \\
\bottomrule
\end{tabular}
\end{table}
\subsection{\ours with different scoring strategies}
A key advantage of enhancing post-hoc methods is their modularity, which enables integration with a variety of scoring methods. To evaluate \ours's flexibility, we pair it with EBO~\cite{liu2020energy}, ViM~\cite{haoqi2022vim} and GEN~\cite{liu2023gen}, and compare the results to their corresponding standard performance. \Cref{tab:modularity} presents the results over the four main OpenOOD dataset scenarios. Applying \ours guarantees a consistent performance improvement, regardless of the scoring method used.
\begin{table}[t]
\centering
\setlength{\tabcolsep}{4pt}
\newcolumntype{G}{@{}p{1.0em}@{}}
\caption{\ours performance with different scoring strategies. ViM* is initialized with
evaluation data given the lack of access to some of the training data.
\textbf{Bold} indicates best performance.}
\resizebox{\linewidth}{!}{%
\begin{tabular}{l  c  c  c  G  c  c  c  G  c  c  c  G  c  c  c}
\toprule
  & \multicolumn{3}{c}{\textbf{CIFAR-10}}     &
  & \multicolumn{3}{c}{\textbf{CIFAR-100}}    &
  & \multicolumn{3}{c}{\textbf{ImageNet-200}} &
  & \multicolumn{3}{c}{\textbf{ImageNet}}     \\
\cmidrule(lr){2-4}   \cmidrule(lr){6-8}
\cmidrule(lr){10-12} \cmidrule(lr){14-16}
Method
  & AUC$\uparrow$ & FPR$\downarrow$ & AUPR$\uparrow$ &
  & AUC$\uparrow$ & FPR$\downarrow$ & AUPR$\uparrow$ &
  & AUC$\uparrow$ & FPR$\downarrow$ & AUPR$\uparrow$ &
  & AUC$\uparrow$ & FPR$\downarrow$ & AUPR$\uparrow$ \\
\midrule
EBO
  & 90.00 & 48.24 & 91.88 &
  & 80.15 & 56.26 & 81.85 &
  & 87.51 & 45.01 & 88.11 &
  & 84.03 & 50.46 & 58.54 \\
\rowcolor{gray!20}
EBO + \textbf{\ours}
  & $\mathbf{90.24}$ & $\mathbf{40.16}$ & $\mathbf{92.03}$ &
  & $\mathbf{82.09}$ & $\mathbf{52.31}$ & $\mathbf{83.98}$ &
  & $\mathbf{89.51}$ & $\mathbf{36.70}$ & $\mathbf{90.06}$ &
  & $\mathbf{86.55}$ & $\mathbf{40.92}$ & $\mathbf{64.03}$ \\
\midrule
ViM*
  & 90.61 & 34.31 & 91.94 &
  & 75.26 & 62.69 & $\mathbf{78.57}$ &
  & 82.86 & $\mathbf{48.39}$ & 83.59 &
  & $\mathbf{84.29}$ & $\mathbf{43.04}$ & $\mathbf{61.44}$ \\
\rowcolor{gray!20}
ViM* + \textbf{\ours}
  & $\mathbf{91.12}$ & $\mathbf{33.54}$ & $\mathbf{92.62}$ &
  & $\mathbf{75.28}$ & $\mathbf{61.41}$ & 78.48 &
  & $\mathbf{83.47}$ & 48.51 & $\mathbf{85.18}$ &
  & 82.47 & 45.13 & 56.95 \\
\midrule
GEN
  & 90.30 & 41.04 & 91.76 &
  & 80.23 & 55.95 & 81.94 &
  & 88.29 & 41.34 & 88.58 &
  & 84.60 & 47.49 & 57.60 \\
\rowcolor{gray!20}
GEN + \textbf{\ours}
  & $\mathbf{90.61}$ & $\mathbf{37.43}$ & $\mathbf{92.45}$ &
  & $\mathbf{82.47}$ & $\mathbf{51.88}$ & $\mathbf{84.65}$ &
  & $\mathbf{89.03}$ & $\mathbf{37.91}$ & $\mathbf{89.69}$ &
  & $\mathbf{85.89}$ & $\mathbf{42.80}$ & $\mathbf{60.74}$ \\
\bottomrule
\end{tabular}
}
\label{tab:modularity}
\end{table}

\subsection{Results beyond selecting the penultimate layer representations}

\subsubsection{Comparison with $\ell_2$-normalization.}
\begin{figure}[t]
    \resizebox{0.55\linewidth}{!}{
    \begin{minipage}[c]{.64\linewidth}
        \begin{tabular}{l *{5}{c}}
        \toprule
        & \textbf{C-10} & \textbf{C-100} & \multicolumn{3}{c}{\textbf{IN}} \\
        \cmidrule(lr){2-2} \cmidrule(lr){3-3} \cmidrule(lr){4-6}
        & \textbf{RN18} & \textbf{RN18} & \textbf{RN50} & \textbf{CN-B} & \textbf{SWIN} \\
        MDS & $69.5$ & $79.5$ & $49.5$ & $33.6$ & $52.4$ \\
        MDS++ & $46.2$ & $72.9$ & $52.0$ & $24.3$ & $38.9$ \\
        \rowcolor{gray!20}MDS + \textbf{\ours} & $\mathbf{41.0}$ & $\mathbf{50.0}$ & $\mathbf{48.9}$ & $\mathbf{22.1}$ & $\mathbf{37.6}$ \\
        \midrule
        RMDS & $52.5$ & $76.1$ & $62.5$ & $31.7$ & $48.7$ \\
        RMDS++ & $54.1$ & $77.5$ & $70.4$ & $29.5$ & $39.3$\\
        \rowcolor{gray!20}RMDS + \textbf{\ours} & $\mathbf{47.1}$ & $\mathbf{52.1}$ & $\mathbf{49.8}$ & $\mathbf{23.3}$ & $\mathbf{38.8}$ \\ 
        \bottomrule
        \end{tabular}
        \captionof{table}{
        \ours vs $\ell_2$-normalization. Average FPR on OpenOOD datasets using the checkpoints used in the MDS++ paper~\cite{muller2025mahalanobis}, rather than the OpenOOD defaults. \textbf{Bold} indicates best.
        }
        \label{tab:wt_v_l2}
    \end{minipage}
    }
    \quad
    \begin{minipage}[c]{.38\linewidth}
    \centering \includegraphics[width=\linewidth]{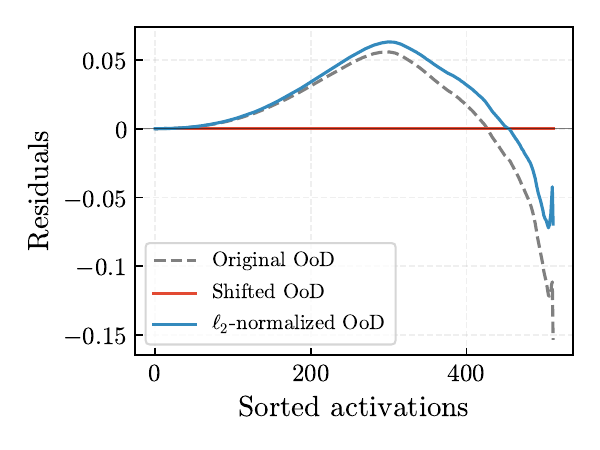}
        \captionof{figure}{
        \ours (shifted) vs $\ell_2$-normalization, shown as residuals with respect to the average ranked ID activations.
        }
        \label{fig:wt_v_l2}
    \end{minipage}
\end{figure}

MDS~\cite{lee2018simple} scores samples by their Mahalanobis distance to the nearest class mean in feature space, while RMDS~\cite{ren2021simple} refines this by computing distances relative to a class-agnostic background distribution, reducing the influence of uninformative feature dimensions. Recent methods, such as MDS++ and RMDS++~\cite{muller2025mahalanobis}, have demonstrated that normalizing activations to match the ID $\ell_2$-norm can significantly bolster OoD detection performance compared to the original unnormalized versions. \ours takes this further by aligning the full activation distribution to the ID reference via $\boldsymbol{\mu}$. This transformation implicitly enforces $\ell_2$-norm matching, since every activation vector is replaced by the same reference vector, but produces a qualitatively different vector geometry from simple radial scaling, as illustrated in~\cref{fig:wt_v_l2}. We show that \ours provides a more robust enhancement to \textsc{MDS} and \textsc{RMDS} than the standard $\ell_2$-normalized variants. Empirical results in~\cref{tab:wt_v_l2} confirm that this activation shifting strategy consistently outperforms standard $\ell_2$-normalization across both baseline frameworks.

\subsubsection{Single layer selection.}
\begin{figure}[t]
    \centering
    \includegraphics[width=\linewidth]{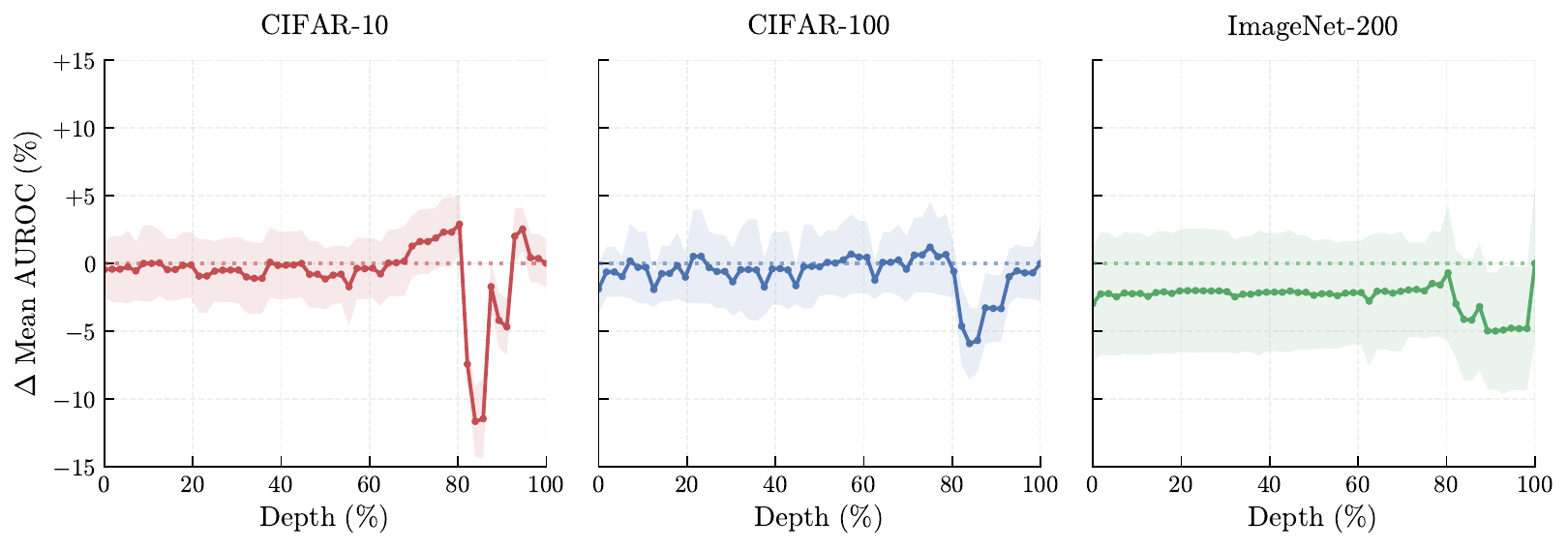}
    \caption{Mean AUROC gain of \ours (solid line) when applied to each ResNet layer, relative to the commonly selected penultimate layer (dashed line).}
    \label{fig:plot_single_layers}
\end{figure}

Our proposed \ours method is applicable to any intermediate model representation, offering flexibility on the layer selection. A natural choice is the penultimate layer of the model, which has been shown to be the most effective for other enhancing post-hoc methods, such as ReAct~\cite{sun2021react} and ASH~\cite{djurisic2023extremely}. Nevertheless, non-penultimate intermediate layers have been proven to be effective for OoD detection~\cite{guglielmo2025leveraging, meza2025mysteries,harun2024variables}. We present results in \cref{fig:plot_single_layers} for the CIFAR-10, CIFAR-100 and ImageNet-200 scenarios, where \ours has been independently applied to each intermediate layer. Across all results, there is only a noticeable improvement over the penultimate layer for the CIFAR-10 scenario, which occurs in a batch normalization layer inside the 4th residual block. In the other cases, the improvement is either not significant or not observed at all, leading to the conclusion that applying \ours to the penultimate layer remains the most straightforward single-layer choice. We explore multiple layer selection in the Supplementary Material.

\subsection{How much ID data is needed?}
\begin{figure}[h]
    \centering
    \begin{minipage}[t]{0.48\linewidth}
        \centering
        \includegraphics[width=\linewidth]{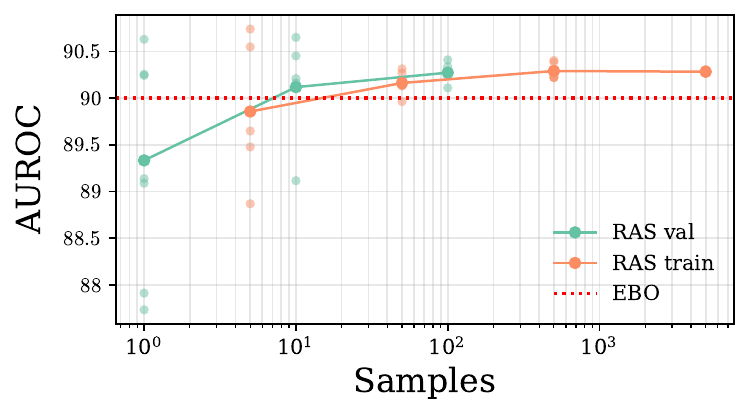}
        \subcaption{CIFAR-10}
    \end{minipage}\hfill
    \begin{minipage}[t]{0.48\linewidth}
        \centering
        \includegraphics[width=\linewidth]{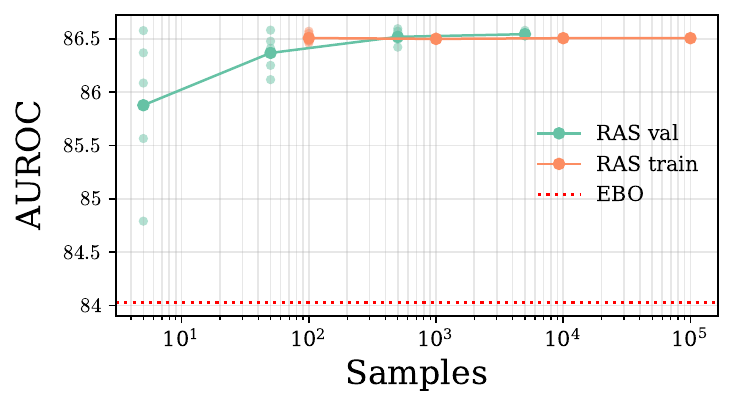}
        \subcaption{ImageNet}
    \end{minipage}
    \caption{AUROC on CIFAR-10 (left) and ImageNet (right) as ID.}
    \label{fig:subsampling}
\end{figure}

Earlier, we computed the reference profile using the entire training set. Here, we evaluate \ours with varying numbers of samples, using both training data and unlabeled held-out validation data, showing that retaining the original training set is also unnecessary. Figure~\ref{fig:subsampling} (log scale) confirms nearly identical results across both splits. Only a few samples are enough to improve over the EBO baseline (red dashed line), and held-out validation data not used for training can also be used.

\section{Discussion}
\label{sec:discussion}
Methods such as ReAct operate under the assumption that OoD samples are characterized by abnormally high activations, and improve detection by suppressing them. It is natural to ask whether the opposite direction of correction is also effective: can pushing below-mean activations upward toward $\boldsymbol{\mu}$ also improve OoD detection? Is the benefit of \ours tied to a single direction of activation modification?

To study this, we define \ours-inhibit and \ours-excite, two extensions of \ours which only shift the ranked activations that are \textit{above} and \textit{below} the ranked ID mean, respectively. We report this ablation in \cref{tab:ablation} with the energy score statistics for each variant relative to the unmodified EBO baseline. Each entry shows both the baseline and the post-transform energy score distribution $\mu \pm \sigma$, along with the change in mean $\Delta\mu$ and the change in AUROC, allowing us to disentangle the respective contributions of \ours-inhibit and \ours-excite to the overall score separation. \Cref{tab:ablation} shows that, crucially, \emph{both} directions improve over EBO in the majority of settings, demonstrating that the benefit of \ours is not tied to a single direction of activation modification. Full \ours compounds both effects simultaneously, achieving the largest average AUROC gain in almost every scenario, up to $+2.45$ and $+2.86$ for ImageNet-200 and ImageNet respectively. Consistently across all benchmarks and variants, \ours also reduces the standard deviation of energy scores for both ID and OoD samples, with the improvement driven primarily by this variance compression rather than any shift in means. 

Therefore, the benefit of activation editing is not tied to any particular direction of correction: both upward and downward shifts independently improve OoD separation, and \ours captures the full gain by applying both.

\tabinhibitexcite 

\subsection{Activation study} 
A further question is whether the improvement holds regardless of the relative behavior of OoD and ID activations across rank positions, or whether it is specific to certain configurations.  To study this, we compare the mean ranked activation profiles of ID and OoD datasets. Let $\bar{a}_{\pi_i}$ denote the mean activation at rank $i$ across a selected OoD dataset, where $\pi$ is the rank-order permutation. We define the signed deviation between the OoD dataset profile and the ID reference at each rank position $i$ as the \emph{residual}:
\begin{equation}
    \Delta \bar{a}_{\pi_i} = \bar{a}_{\pi_i} - \mu_i\,,
\end{equation}

\figresiduals  

\Cref{fig:residuals} shows two qualitatively distinct behaviors when observing the mean ranked profiles and their corresponding residuals for CIFAR-10 and ImageNet as ID datasets. For CIFAR-10, OoD samples exhibit a characteristic profile where high-rank activations fall below the ID reference, while low-rank activations exceed it; while for ImageNet the pattern is reversed. \ours-excite and \ours-inhibit are both effective across all cases, confirming that \ours does not rely on any particular configuration of the OoD activation profile relative to the ID reference. This suggests that the benefit of activation shifting is a general property of the shifting to the ID reference profile, rather than a consequence of any specific OoD behavior. Additional profiles are provided in the Supplementary Material. 

\section{Conclusions}
\label{sec:conclusions}
Our analysis reveals why competing methods such as ASH and SCALE exhibit inconsistent behavior across benchmarks: their underlying assumptions on the $\mu/\sigma$ ratio of the penultimate layer activations are often violated, and their reliance on a ratio of activation sums breaks down entirely in the absence of rectification. \ours sidesteps these failure modes by operating on rank order rather than activation magnitude, and by anchoring each sample to the same ID reference regardless of the direction or magnitude of its deviation.

We present a hyperparameter-free post-hoc method for out-of-distribution detection that shifts penultimate layer activations to match the average ranked activation profile of in-distribution data. Unlike existing enhancing methods, \ours requires no threshold tuning, no access to OoD data, and no assumptions about the sign or distribution of the penultimate layer activations, making it directly applicable to a broad range of architectures, including ReLU-based CNNs, GELU-based ConvNeXts, and vision transformers.

We validate \ours across four standard, well-established benchmarks, multiple architectures, and three scoring functions, consistently matching or outperforming state-of-the-art enhancing methods that operate at their optimized hyperparameter values. Crucially, as shown in the~\cref{tab:tab_acc}, \ours leaves ID classification accuracy virtually unchanged. We further show that the improvement is not tied to a single direction of activation modification: both inhibiting and excitatory shifts toward $\boldsymbol{\mu}$ independently improve OoD separation, with \ours compounding both effects through variance compression.

\section*{Acknowledgments}
Gianluca Guglielmo acknowledges the support of KAI GmbH and Infineon Technologies Austria. This research was partially funded by the Austrian Science Fund (FWF) 10.55776/COE12. The authors would like to thank Christoph Griesbacher for the fruitful discussions during the development of this research. We would also like to thank Thomas Pock for his academic supervision at Graz University of Technology.

\bibliographystyle{splncs04}
\bibliography{ref}

\newpage

\appendix

\renewcommand{\thesection}{\Alph{section}}
\renewcommand{\thetable}{S\arabic{table}}
\renewcommand{\thefigure}{S\arabic{figure}}
\setcounter{section}{0}

\section{Extended Results}

\Crefrange{tab:complete_1}{tab:complete_9} show the complete, in-depth results of our experiments. Each table contains the results per OoD dataset and a final average. For the models available on OpenOOD, we provide a comparison with every post-hoc OoD detection method logged on the official leaderboard. For better readability, the tables have been moved to the end of the Supplementary section.

\subsubsection{Additional methods.}
The following methods appear in the OpenOOD leaderboard but were not covered in the main text. \textit{TempScale}~\cite{liang2018enhancing} calibrates the softmax output via temperature scaling before computing the MSP score. \textit{KNN}~\cite{sun2022out} scores a test sample by its distance to the $k$-th ID nearest neighbor in the feature space. \textit{RankFeat}~\cite{song2022rankfeat} removes the rank-1 component of the intermediate feature map via SVD. \textit{GradNorm}~\cite{huang2021importance} computes the $\ell_1$ norm of the gradients of the KL divergence between the softmax output and a uniform distribution, requiring a full backward pass at inference. \textit{MDSEns}~\cite{lee2018simple} aggregates Mahalanobis distance scores computed at multiple layers. \textit{MLS}~\cite{hendrycks2022scaling} uses the maximum raw logit as the OOD score. \textit{KLM}~\cite{hendrycks2022scaling} scores samples using the KL divergence between their softmax output and the per-class softmax distributions estimated from ID data. \textit{OpenMax}~\cite{bendale2016towards} extends the SoftMax function to include another open-set class that encodes unknowns. \textit{Gram}~\cite{sastry2020detecting} measures deviations in the Gram matrix statistics of intermediate feature maps relative to ID data. \textit{SHE}~\cite{zhang2022out} derives an OOD score from a simplified Hopfield energy applied to penultimate-layer activations. 

\subsubsection{Comparison with AdaSCALE.}AdaSCALE~\cite{regmi2025adascale} extends SCALE~\cite{xu2024scaling} by making the scaling percentile adaptive per sample: it estimates a sample's OoD-ness from the activation shift produced by a small input perturbation, which scales likely in-distribution samples more strongly and likely OoD ones less. Estimating this shift requires an extra backward pass per sample, so AdaSCALE does not fit cleanly among the single-pass score-enhancing methods we compare against in~\cref{tab:full_comparison_table}, therefore we report it separately here. Its activation study is compelling because it shows that, under a small perturbation, OoD inputs shift their high-magnitude activations far more than ID inputs do, giving a simple and measurable signal of OoD-ness. The framing of shifting differs from ours, however: AdaSCALE defines the shift as the difference between the ranked activation profiles before and after the perturbation, whereas we define shifting as the process of ``matching'' the in-distribution reference profile. \Cref{tab:adascale} reports the comparison against both AdaSCALE variants: RAS attains the lowest FPR@95 in six of the eight settings, including every ConvNeXt and ViT backbone, while remaining single-pass and hyperparameter-free, whereas the AdaSCALE variants tend to lead on AUROC.

\begin{table}[h]
\centering
\caption{Comparison against AdaSCALE on the OpenOOD benchmark, reporting AUROC (AUC$\uparrow$) and FPR@95 (FPR$\downarrow$). Best per column in \textbf{bold}, second-best \underline{underlined}; EBO is the base score.}
\resizebox{\linewidth}{!}{
\begin{tabular}{l cc cc cc cc cc cc cc cc}
\toprule
& \multicolumn{2}{c}{\textbf{CIFAR-10}} & \multicolumn{2}{c}{\textbf{CIFAR-100}} & \multicolumn{2}{c}{\textbf{IN-200}} & \multicolumn{10}{c}{\textbf{ImageNet}} \\
\cmidrule(lr){2-3} \cmidrule(lr){4-5} \cmidrule(lr){6-7} \cmidrule(lr){8-17}

& \multicolumn{2}{c}{\textbf{ResNet18}} & \multicolumn{2}{c}{\textbf{ResNet18}} & \multicolumn{2}{c}{\textbf{ResNet18}} & \multicolumn{2}{c}{\textbf{ResNet50}} & \multicolumn{2}{c}{\textbf{EffNet-B0}} & \multicolumn{2}{c}{\textbf{ConvNx-T}} & \multicolumn{2}{c}{\textbf{ConvNx-B}} & \multicolumn{2}{c}{\textbf{ViT-B/16}} \\

\cmidrule(lr){2-3} \cmidrule(lr){4-5} \cmidrule(lr){6-7} \cmidrule(lr){8-9} \cmidrule(lr){10-11} \cmidrule(lr){12-13} \cmidrule(lr){14-15} \cmidrule(lr){16-17} & AUC$\uparrow$ & FPR$\downarrow$ & AUC$\uparrow$ & FPR$\downarrow$ & AUC$\uparrow$ & FPR$\downarrow$ & AUC$\uparrow$ & FPR$\downarrow$ & AUC$\uparrow$ & FPR$\downarrow$ & AUC$\uparrow$ & FPR$\downarrow$ & AUC$\uparrow$ & FPR$\downarrow$ & AUC$\uparrow$ & FPR$\downarrow$ \\
\midrule
EBO & $\underline{90.00}$ & $\underline{48.24}$ & $80.15$ & $56.26$ & $87.51$ & $45.01$ & $84.03$ & $50.46$ & $76.76$ & $72.29$ & $69.36$ & $83.03$ & $59.36$ & $93.29$ & $72.35$ & $88.48$ \\
\midrule
AdaSCALE-A & $86.70$ & $65.91$ & $81.58$ & $54.04$ & $\underline{90.82}$ & $\underline{36.45}$ & $\mathbf{91.62}$ & $\mathbf{29.95}$ & $\mathbf{84.62}$ & $\underline{50.40}$ & $\mathbf{84.32}$ & $50.71$ & $\underline{85.79}$ & $47.18$ & $\mathbf{82.43}$ & $56.73$ \\
AdaSCALE-L & $77.76$ & $85.03$ & $\mathbf{82.43}$ & $\underline{53.18}$ & $\mathbf{91.16}$ & $\mathbf{35.00}$ & $\underline{91.44}$ & $\underline{30.68}$ & $\underline{84.15}$ & $50.47$ & $\underline{83.58}$ & $\underline{50.63}$ & $\mathbf{85.97}$ & $\underline{46.48}$ & $\underline{82.39}$ & $\underline{56.63}$ \\
\rowcolor{gray!20}\textbf{\ours} & $\mathbf{90.24}$ & $\mathbf{40.16}$ & $\underline{82.09}$ & $\mathbf{52.31}$ & $89.51$ & $36.70$ & $86.55$ & $40.92$ & $83.78$ & $\mathbf{46.87}$ & $83.31$ & $\mathbf{46.28}$ & $84.86$ & $\mathbf{45.29}$ & $81.48$ & $\mathbf{55.19}$ \\
\bottomrule
\end{tabular}
}
\label{tab:adascale}
\end{table}

\subsubsection{Multiple layer selection.}
To study the effect of shifting multiple hidden layers simultaneously, we also included a variant of our method called \ours HL. The choice of layers follows the natural modular structure of each architecture. For the ResNet-based architectures, we select all residual blocks; for EfficientNet, all MBConv blocks; for ViT, all transformer encoder blocks; and for ConvNeXt, all ConvNeXt blocks. 

Overall, \ours HL performs comparably to \ours. While it occasionally outperforms the original, notably on CIFAR-10 and CIFAR-100 with ResNet18 (\crefrange{tab:complete_1}{tab:complete_2}), it falls short in other settings, such as ImageNet with ResNet50 and ConvNeXt-Tiny (\crefrange{tab:complete_4}{tab:complete_5}). The added complexity of ranking multiple layers is therefore not justified by the marginal gains.

\subsubsection{Class-conditional reference profile.}
A simple extension to RAS is to build multiple reference profiles, one for each known class. We test it by first predicting the label without RAS, then applying RAS using the pre-computed class-specific reference profile. We select ImageNet to test this hypothesis, as it contains \num{1000} highly diverse classes.~\Cref{tab:cc} shows that class-conditioning does not yield gains over the standard unconditional RAS.
\begin{table}[h]
    \centering
    \caption{AUROC for ImageNet as ID. CC: class-conditioned.}
    \label{tab:cc}
    \begin{tabular}{lcccccc}
    \toprule
        & \textbf{SSB} & \textbf{NINCO} & \textbf{iNat} & \textbf{DTD} & \textbf{OpnImg} & \textbf{AVG.}  \\
    \midrule
        RAS & $\mathbf{71.00}$ & $\mathbf{82.50}$ & $\mathbf{95.23}$ & $\mathbf{92.30}$ & $\mathbf{91.73}$ & $\mathbf{86.55}$ \\
        RAS-CC & $70.29$ & $79.22$ & $92.74$ & $90.80$ & $88.92$ & $84.39$ \\
    \bottomrule
    \end{tabular}
\end{table}

\newpage
\section{Covariate Shift}
\begin{figure}[h!]
    \centering
    \includegraphics[width=1\linewidth]{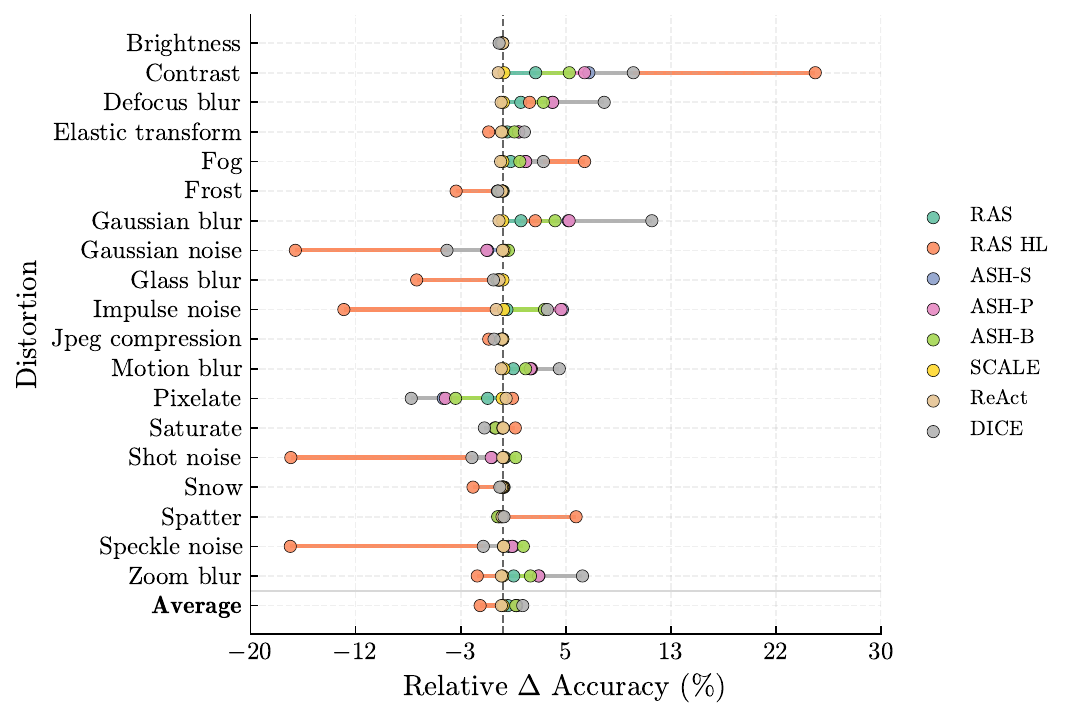}
    \caption{Relative classification accuracy differences with respect to the unmodified model, for different CIFAR-10 corruptions. All scoring enhancing methods on average keep accuracy stable. \ours HL has strong improvements on some corruptions, such as \texttt{contrast}, but these are balanced by a detrimental performance on other corruptions.}
    \label{fig:plot_corruptions}
\end{figure}

Beyond semantic shifts, robust OoD detection must also handle inputs affected by covariate shift. CIFAR-10C~\cite{hendrycksbenchmarking} benchmarks this setting by applying a range of corruption types and severities to CIFAR-10 samples. We evaluate how activation shifting affects classification accuracy under covariate shifts by applying \ours and comparable enhancing methods to CIFAR-10C. As shown in \cref{fig:plot_corruptions}, all methods preserve accuracy to the same degree as the unmodified baseline.

Full-spectrum OoD~(FSOOD)~\cite{yang2023full} is an OoD benchmark that includes covariate-shifted inputs as ID, alongside the standard ones. As with the normal task, the aim is to correctly-classify all ID inputs and, concurrently, to distinguish them from OoD samples, with the additional difficulty coming from the covariate-shifted inputs confounding the model. \Crefrange{tab:fsood_in200}{tab:fsood_in} show the results for ImageNet-200 and ImageNet as ID. \ours is comparable to the other score-enhancing methods, apart from ASH(-S), which is the best performing one.

\begin{table}[h!]
\centering
\caption{FSOOD - ImageNet200}
\label{tab:fsood_in200}
\resizebox{\linewidth}{!}{
\begin{tabular}{l c c c c c c c c c c c c c c c c c c}
\toprule
 & \multicolumn{3}{c}{\textbf{SSB\_HARD}} & \multicolumn{3}{c}{\textbf{NINCO}} & \multicolumn{3}{c}{\textbf{INATURALIST}} & \multicolumn{3}{c}{\textbf{TEXTURES}} & \multicolumn{3}{c}{\textbf{OPENIMAGE\_O}} & \multicolumn{3}{c}{\textbf{AVERAGE}} \\
Method & AUROC & FPR & AUPR & AUROC & FPR & AUPR & AUROC & FPR & AUPR & AUROC & FPR & AUPR & AUROC & FPR & AUPR & AUROC & FPR & AUPR \\
\midrule
OpenMax & $47.64$ & $91.55$ & $50.93$ & $54.15$ & $85.00$ & $13.08$ & $72.44$ & $68.19$ & $33.91$ & $69.12$ & $76.72$ & $18.00$ & $62.31$ & $77.39$ & $34.09$ & $61.13$ & $79.77$ & $30.00$ \\
MSP & $50.94$ & $\underline{89.08}$ & $51.41$ & $57.76$ & $\underline{79.30}$ & $13.24$ & $70.42$ & $67.24$ & $30.11$ & $65.11$ & $79.79$ & $18.90$ & $62.80$ & $74.22$ & $33.38$ & $61.41$ & $77.93$ & $29.41$ \\
TempScale & $50.05$ & $\underline{89.33}$ & $50.83$ & $56.86$ & $\underline{79.51}$ & $12.95$ & $70.18$ & $66.61$ & $29.64$ & $65.65$ & $79.67$ & $19.46$ & $62.29$ & $74.10$ & $33.03$ & $61.01$ & $77.84$ & $29.18$ \\
ODIN & $44.31$ & $91.71$ & $46.50$ & $52.36$ & $86.66$ & $11.63$ & $70.19$ & $65.50$ & $28.12$ & $67.10$ & $79.13$ & $19.90$ & $61.48$ & $76.03$ & $31.55$ & $59.09$ & $79.81$ & $27.54$ \\
MDS & $48.59$ & $93.90$ & $51.88$ & $56.65$ & $88.52$ & $14.22$ & $68.25$ & $74.46$ & $29.79$ & $73.84$ & $74.10$ & $29.33$ & $61.90$ & $83.68$ & $35.53$ & $61.85$ & $82.93$ & $32.15$ \\
MDSEns & $34.22$ & $96.28$ & $41.87$ & $41.58$ & $95.83$ & $9.46$ & $43.63$ & $91.39$ & $15.30$ & $67.54$ & $84.83$ & $25.86$ & $48.38$ & $93.52$ & $25.23$ & $47.07$ & $92.37$ & $23.54$ \\
RMDS & $\underline{56.24}$ & $89.44$ & $\underline{57.49}$ & $\underline{60.95}$ & $\mathbf{79.23}$ & $\underline{15.34}$ & $71.71$ & $65.62$ & $29.96$ & $64.61$ & $76.73$ & $16.87$ & $63.52$ & $74.82$ & $34.35$ & $63.41$ & $77.17$ & $30.80$ \\
Gram & $\mathbf{59.12}$ & $\mathbf{86.37}$ & $\mathbf{58.06}$ & $\mathbf{63.35}$ & $87.59$ & $\mathbf{19.31}$ & $58.42$ & $86.24$ & $21.56$ & $75.86$ & $82.76$ & $36.61$ & $61.51$ & $87.23$ & $36.98$ & $63.65$ & $86.04$ & $34.50$ \\
EBO & $47.56$ & $90.71$ & $49.39$ & $53.45$ & $83.61$ & $12.08$ & $65.91$ & $70.53$ & $25.00$ & $68.03$ & $79.46$ & $22.35$ & $59.83$ & $77.14$ & $31.43$ & $58.96$ & $80.29$ & $28.05$ \\
OpenGAN & $41.58$ & $95.90$ & $47.78$ & $46.51$ & $94.47$ & $11.04$ & $61.78$ & $81.85$ & $26.85$ & $59.02$ & $84.76$ & $14.19$ & $58.53$ & $86.71$ & $32.76$ & $53.48$ & $88.74$ & $26.52$ \\
GradNorm & $53.50$ & $91.16$ & $54.35$ & $55.32$ & $91.68$ & $13.43$ & $70.38$ & $78.68$ & $32.57$ & $73.12$ & $82.11$ & $33.36$ & $63.52$ & $84.81$ & $37.32$ & $63.17$ & $85.69$ & $34.21$ \\
ReAct & $47.25$ & $91.22$ & $49.48$ & $53.84$ & $84.61$ & $12.33$ & $69.45$ & $67.52$ & $28.06$ & $71.45$ & $72.82$ & $24.46$ & $62.30$ & $74.81$ & $33.34$ & $60.86$ & $78.20$ & $29.53$ \\
MLS & $48.18$ & $90.68$ & $49.82$ & $54.43$ & $83.27$ & $12.34$ & $67.94$ & $69.50$ & $27.41$ & $67.12$ & $79.39$ & $20.91$ & $60.66$ & $76.55$ & $32.17$ & $59.67$ & $79.88$ & $28.53$ \\
KLM & $\underline{54.03}$ & $91.97$ & $\underline{57.36}$ & $\underline{60.52}$ & $85.33$ & $\underline{15.72}$ & $75.37$ & $66.87$ & $\underline{37.54}$ & $64.12$ & $80.11$ & $15.54$ & $\underline{66.20}$ & $77.94$ & $37.68$ & $64.05$ & $80.44$ & $32.77$ \\
VIM & $45.34$ & $91.61$ & $48.87$ & $57.09$ & $82.35$ & $13.85$ & $71.34$ & $68.15$ & $33.01$ & $\mathbf{82.54}$ & $\mathbf{58.50}$ & $\underline{42.10}$ & $\underline{65.70}$ & $74.54$ & $\underline{38.54}$ & $\underline{64.40}$ & $\underline{75.03}$ & $35.27$ \\
KNN & $44.05$ & $91.73$ & $47.70$ & $54.51$ & $81.23$ & $12.54$ & $71.53$ & $69.10$ & $33.26$ & $\underline{81.88}$ & $\underline{69.06}$ & $\mathbf{49.30}$ & $62.12$ & $74.43$ & $36.49$ & $62.82$ & $77.11$ & $\underline{35.86}$ \\
DICE & $48.11$ & $90.94$ & $50.61$ & $54.31$ & $84.24$ & $12.72$ & $66.53$ & $72.10$ & $27.66$ & $72.51$ & $78.84$ & $34.85$ & $61.73$ & $77.79$ & $35.39$ & $60.64$ & $80.78$ & $32.25$ \\
RankFeat & $46.41$ & $95.78$ & $50.58$ & $43.03$ & $96.98$ & $10.33$ & $22.74$ & $99.16$ & $11.76$ & $20.60$ & $99.75$ & $6.36$ & $40.74$ & $98.16$ & $22.89$ & $34.70$ & $97.97$ & $20.38$ \\
ASH & $50.96$ & $90.29$ & $52.35$ & $58.51$ & $84.21$ & $14.53$ & $\mathbf{77.96}$ & $\underline{63.16}$ & $\mathbf{45.46}$ & $\underline{79.39}$ & $\underline{65.99}$ & $\underline{42.28}$ & $\mathbf{69.09}$ & $\underline{71.69}$ & $\mathbf{42.94}$ & $\mathbf{67.18}$ & $\underline{75.07}$ & $\mathbf{39.51}$ \\
SHE & $52.82$ & $91.16$ & $55.08$ & $56.64$ & $86.49$ & $13.78$ & $72.20$ & $71.48$ & $35.98$ & $74.27$ & $78.98$ & $36.26$ & $64.95$ & $79.50$ & $\underline{39.33}$ & $\underline{64.18}$ & $81.52$ & $\underline{36.09}$ \\
GEN & $48.33$ & $89.59$ & $49.75$ & $54.85$ & $80.12$ & $12.37$ & $68.94$ & $66.47$ & $29.29$ & $66.58$ & $79.30$ & $20.72$ & $60.87$ & $73.96$ & $32.51$ & $59.91$ & $77.89$ & $28.93$ \\
\rowcolor{gray!20} \textbf{\ours} & $47.42$ & $90.84$ & $49.33$ & $56.56$ & $80.03$ & $13.07$ & $\underline{77.04}$ & $\mathbf{60.10}$ & $\underline{38.31}$ & $72.30$ & $70.65$ & $23.25$ & $64.71$ & $\mathbf{70.74}$ & $34.94$ & $63.61$ & $\mathbf{74.47}$ & $31.78$ \\
\rowcolor{gray!20} \textbf{\ours} HL & $47.81$ & $90.86$ & $49.76$ & $56.90$ & $80.56$ & $13.31$ & $\underline{76.14}$ & $\underline{61.03}$ & $37.14$ & $69.97$ & $72.59$ & $20.03$ & $64.46$ & $\underline{71.16}$ & $34.49$ & $63.06$ & $75.24$ & $30.95$ \\
\bottomrule
\end{tabular}
}
\end{table}

\begin{table}[h!]
\centering
\caption{FSOOD - ImageNet}
\label{tab:fsood_in}
\resizebox{\linewidth}{!}{
\begin{tabular}{l c c c c c c c c c c c c c c c c c c}
\toprule
 & \multicolumn{3}{c}{\textbf{SSB\_HARD}} & \multicolumn{3}{c}{\textbf{NINCO}} & \multicolumn{3}{c}{\textbf{INATURALIST}} & \multicolumn{3}{c}{\textbf{TEXTURES}} & \multicolumn{3}{c}{\textbf{OPENIMAGE\_O}} & \multicolumn{3}{c}{\textbf{AVERAGE}} \\
Method & AUROC & FPR & AUPR & AUROC & FPR & AUPR & AUROC & FPR & AUPR & AUROC & FPR & AUPR & AUROC & FPR & AUPR & AUROC & FPR & AUPR \\
\midrule
OpenMax & $53.79$ & $86.76$ & $36.07$ & $60.28$ & $76.41$ & $7.19$ & $80.30$ & $49.98$ & $25.64$ & $73.54$ & $62.28$ & $9.84$ & $71.88$ & $60.05$ & $23.22$ & $67.96$ & $67.10$ & $20.39$ \\
MSP & $56.66$ & $\underline{84.47}$ & $35.47$ & $64.93$ & $72.37$ & $7.83$ & $76.35$ & $61.75$ & $20.90$ & $69.33$ & $75.27$ & $9.57$ & $71.28$ & $67.29$ & $23.64$ & $67.71$ & $72.23$ & $19.48$ \\
TempScale & $55.71$ & $\mathbf{84.23}$ & $34.60$ & $64.60$ & $71.55$ & $7.61$ & $77.33$ & $58.16$ & $20.64$ & $70.84$ & $72.87$ & $10.03$ & $72.32$ & $64.35$ & $23.83$ & $68.16$ & $70.23$ & $19.34$ \\
ODIN & $54.22$ & $86.01$ & $33.63$ & $60.59$ & $80.40$ & $6.91$ & $77.43$ & $57.10$ & $19.81$ & $76.04$ & $67.34$ & $12.43$ & $73.40$ & $65.45$ & $24.80$ & $68.34$ & $71.26$ & $19.52$ \\
MDS & $39.22$ & $95.41$ & $27.44$ & $52.83$ & $86.89$ & $5.81$ & $54.06$ & $83.38$ & $9.86$ & $86.26$ & $55.95$ & $\underline{38.55}$ & $60.75$ & $82.16$ & $19.35$ & $58.62$ & $80.76$ & $20.20$ \\
MDSEns & $37.13$ & $96.22$ & $26.14$ & $47.80$ & $93.56$ & $5.14$ & $53.32$ & $87.47$ & $9.26$ & $73.39$ & $78.56$ & $15.05$ & $53.24$ & $92.72$ & $14.75$ & $52.98$ & $89.71$ & $14.07$ \\
RMDS & $56.61$ & $87.09$ & $37.46$ & $\mathbf{67.50}$ & $\mathbf{70.28}$ & $\underline{8.82}$ & $73.48$ & $55.68$ & $16.41$ & $74.25$ & $67.83$ & $16.92$ & $72.13$ & $61.30$ & $24.32$ & $68.79$ & $68.44$ & $20.79$ \\
Gram & $51.93$ & $90.87$ & $33.29$ & $60.63$ & $86.24$ & $7.67$ & $71.36$ & $72.69$ & $17.33$ & $84.83$ & $64.65$ & $36.83$ & $69.40$ & $79.05$ & $26.11$ & $67.63$ & $78.70$ & $24.25$ \\
EBO & $52.93$ & $86.19$ & $32.93$ & $60.28$ & $76.15$ & $6.68$ & $74.01$ & $55.18$ & $16.43$ & $73.89$ & $66.05$ & $12.09$ & $72.22$ & $60.55$ & $23.52$ & $66.67$ & $68.82$ & $18.33$ \\
GradNorm & $\mathbf{61.33}$ & $\underline{84.44}$ & $\mathbf{40.04}$ & $64.06$ & $85.45$ & $8.69$ & $\underline{87.62}$ & $45.63$ & $\underline{39.57}$ & $85.99$ & $55.84$ & $29.30$ & $76.85$ & $76.76$ & $\underline{36.10}$ & $\underline{75.17}$ & $69.62$ & $\underline{30.74}$ \\
ReAct & $55.34$ & $86.90$ & $36.09$ & $64.51$ & $72.97$ & $8.49$ & $\underline{87.93}$ & $\underline{40.64}$ & $\underline{39.60}$ & $81.08$ & $53.47$ & $19.39$ & $\underline{79.34}$ & $55.89$ & $\underline{35.97}$ & $73.64$ & $61.97$ & $27.91$ \\
MLS & $53.56$ & $85.98$ & $33.32$ & $61.43$ & $75.43$ & $6.92$ & $75.61$ & $54.59$ & $18.04$ & $73.42$ & $66.34$ & $11.26$ & $72.66$ & $60.39$ & $23.75$ & $67.34$ & $68.55$ & $18.66$ \\
KLM & $\underline{56.87}$ & $90.19$ & $\underline{38.91}$ & $\underline{67.26}$ & $74.14$ & $\mathbf{9.22}$ & $80.88$ & $58.60$ & $27.91$ & $70.73$ & $68.69$ & $8.98$ & $74.74$ & $66.22$ & $27.82$ & $70.10$ & $71.57$ & $22.57$ \\
VIM & $45.88$ & $88.85$ & $29.35$ & $59.12$ & $77.85$ & $6.61$ & $72.22$ & $55.59$ & $16.00$ & $\mathbf{93.09}$ & $\mathbf{33.64}$ & $\mathbf{55.92}$ & $75.01$ & $57.23$ & $28.47$ & $69.06$ & $62.63$ & $27.27$ \\
KNN & $43.78$ & $90.50$ & $28.49$ & $59.86$ & $74.77$ & $6.62$ & $67.79$ & $62.46$ & $13.53$ & $\underline{90.29}$ & $\underline{42.59}$ & $\underline{40.79}$ & $69.98$ & $65.03$ & $23.61$ & $66.34$ & $67.07$ & $22.61$ \\
DICE & $54.01$ & $86.93$ & $34.72$ & $60.29$ & $79.87$ & $7.33$ & $82.52$ & $54.89$ & $29.47$ & $83.89$ & $63.84$ & $28.60$ & $76.42$ & $66.53$ & $33.37$ & $71.43$ & $70.41$ & $26.70$ \\
RankFeat & $50.30$ & $93.17$ & $34.98$ & $40.37$ & $96.21$ & $4.67$ & $34.34$ & $96.43$ & $6.82$ & $66.29$ & $83.49$ & $11.92$ & $44.98$ & $93.59$ & $13.21$ & $47.26$ & $92.58$ & $14.32$ \\
ASH & $54.66$ & $84.49$ & $34.93$ & $\underline{66.38}$ & $\underline{70.74}$ & $\underline{8.73}$ & $\mathbf{89.23}$ & $\mathbf{36.82}$ & $\mathbf{43.79}$ & $\underline{89.53}$ & $\underline{38.34}$ & $34.78$ & $\mathbf{81.47}$ & $\mathbf{52.33}$ & $\mathbf{39.33}$ & $\mathbf{76.25}$ & $\mathbf{56.54}$ & $\mathbf{32.31}$ \\
SHE & $\underline{58.15}$ & $85.10$ & $\underline{37.49}$ & $64.27$ & $80.47$ & $8.44$ & $84.71$ & $50.97$ & $32.37$ & $87.48$ & $52.17$ & $33.56$ & $\underline{76.92}$ & $69.38$ & $34.10$ & $\underline{74.31}$ & $67.62$ & $\underline{29.19}$ \\
GEN & $52.95$ & $85.72$ & $33.13$ & $62.73$ & $72.38$ & $7.14$ & $78.47$ & $50.61$ & $24.53$ & $71.82$ & $66.40$ & $9.61$ & $72.62$ & $57.67$ & $23.80$ & $67.72$ & $66.56$ & $19.64$ \\
\rowcolor{gray!20} \textbf{\ours} & $51.45$ & $87.18$ & $32.26$ & $63.17$ & $70.97$ & $7.32$ & $82.69$ & $\underline{42.11}$ & $25.68$ & $78.09$ & $54.23$ & $12.80$ & $75.85$ & $\underline{53.36}$ & $26.31$ & $70.25$ & $\underline{61.57}$ & $20.87$ \\
\rowcolor{gray!20} \textbf{\ours} HL & $52.42$ & $86.92$ & $32.98$ & $64.06$ & $\underline{70.48}$ & $7.58$ & $82.26$ & $42.47$ & $24.66$ & $77.36$ & $54.44$ & $11.69$ & $75.63$ & $\underline{54.16}$ & $25.88$ & $70.35$ & $\underline{61.69}$ & $20.56$ \\
\bottomrule
\end{tabular}
}
\end{table}

\newpage
\section{Profile plots}
\begin{figure}
    \centering
    \includegraphics[width=\linewidth]{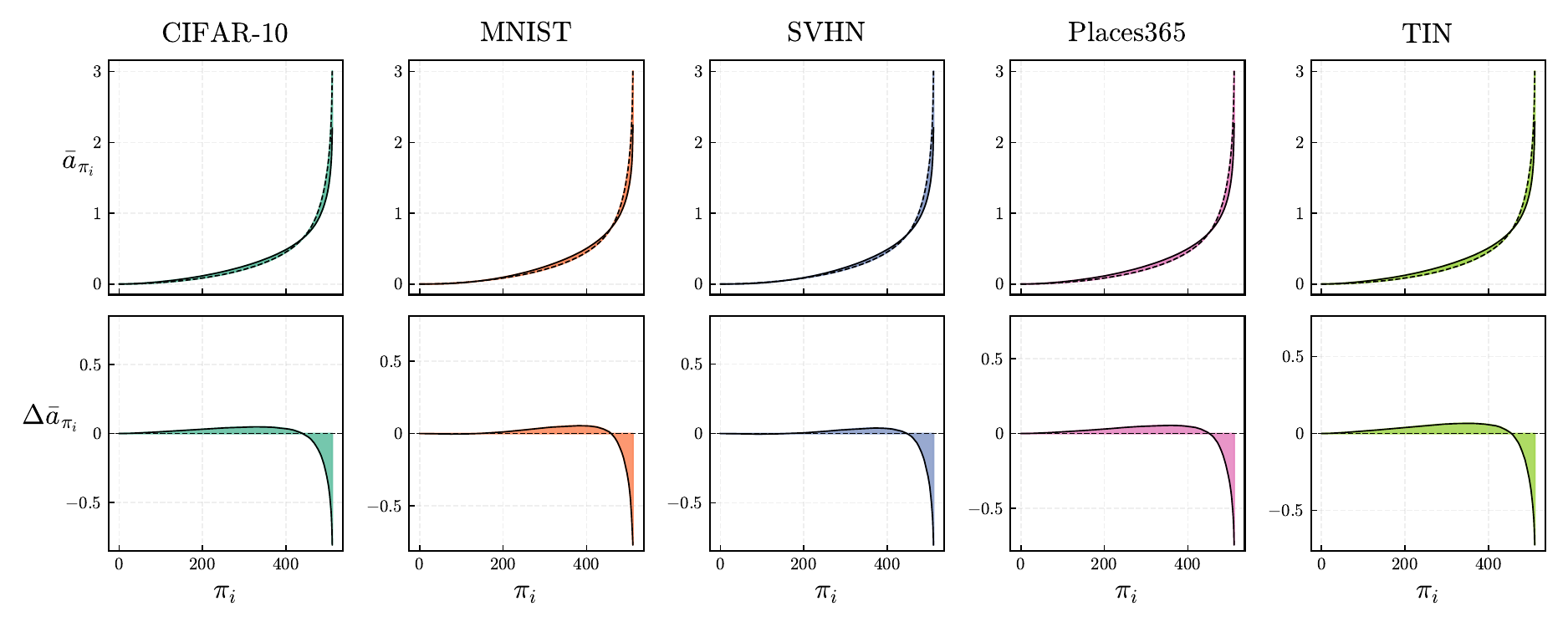}
    \caption{ID dataset: CIFAR-100. Model: ResNet-18.}
    \label{fig:profile_1}
\end{figure}

\begin{figure}
    \centering
    \includegraphics[width=\linewidth]{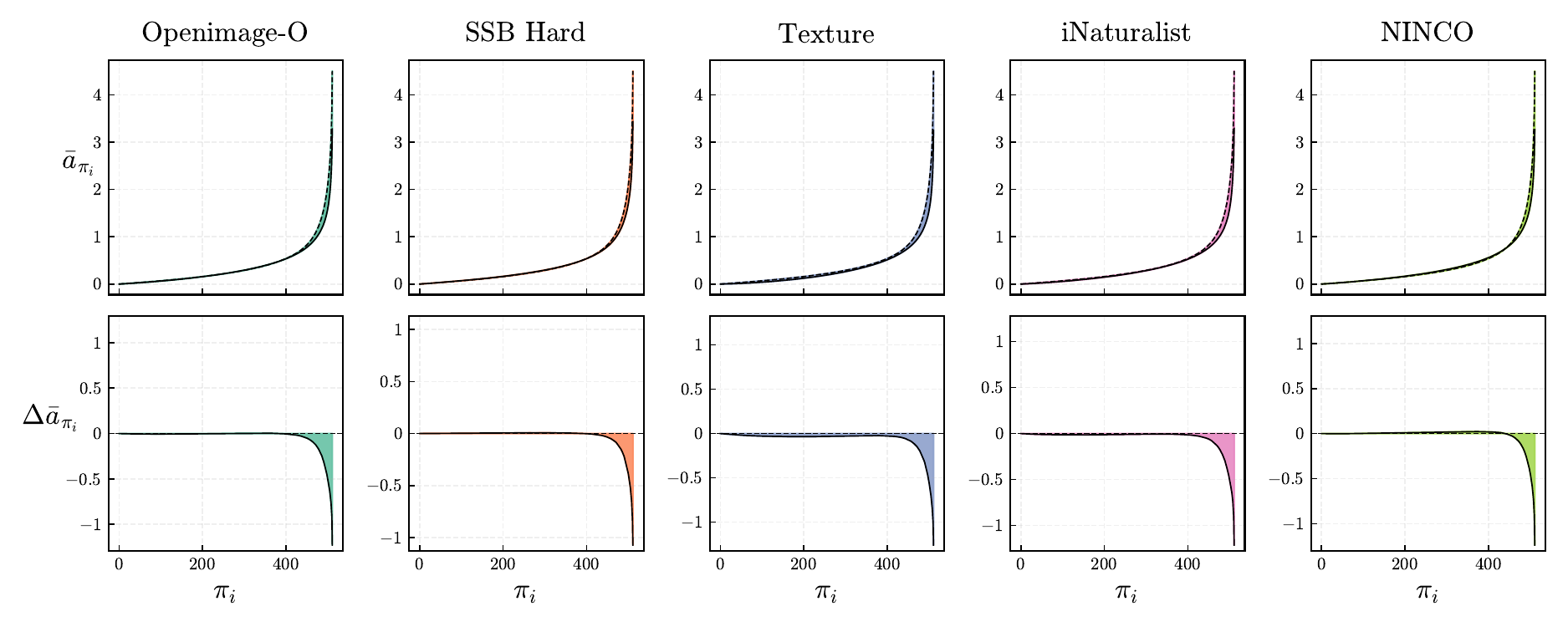}
    \caption{ID dataset: ImageNet-200. Model: ResNet-18.}
    \label{fig:profile_2}
\end{figure}

\begin{figure}
    \centering
    \includegraphics[width=\linewidth]{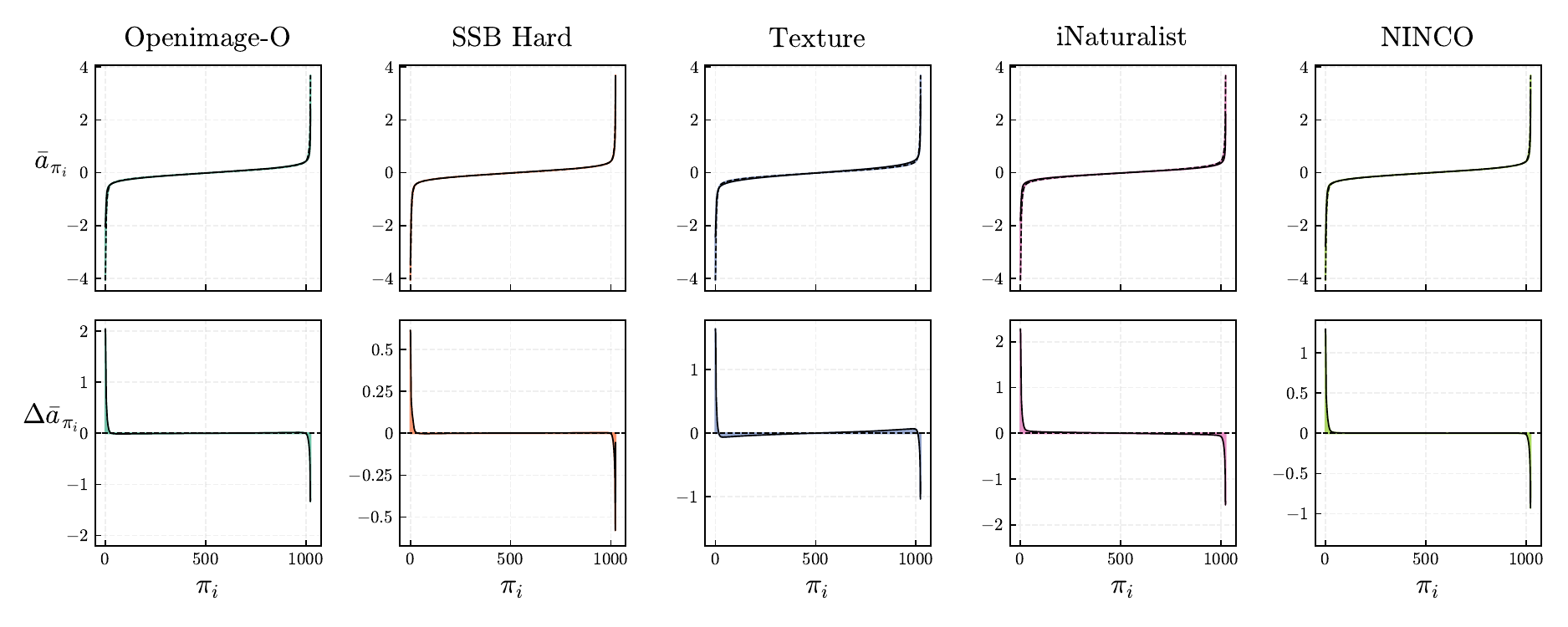}
    \caption{ID dataset: ImageNet. Model: ConvNeXt-Base.}
    \label{fig:profile_3}
\end{figure}

\begin{figure}
    \centering
    \includegraphics[width=\linewidth]{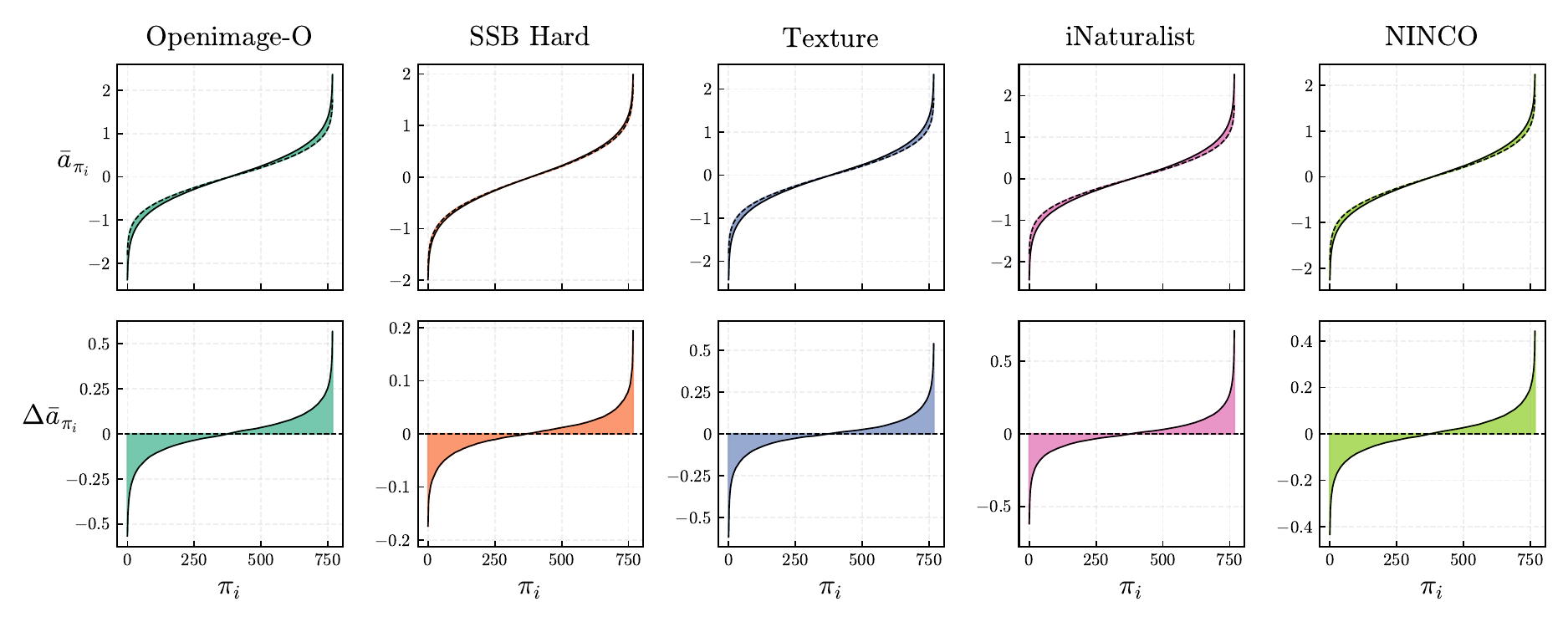}
    \caption{ID dataset: ImageNet. Model: ViT-B/16.}
    \label{fig:profile_4}
\end{figure}

\Crefrange{fig:profile_1}{fig:profile_4} show additional ID/OoD activation profile comparisons. As in~\cref{fig:residuals}, ID is represented by a dashed line, while OoD datasets by solid lines. $\bar{a}_{\pi_i}$ shows the average ranked activations, while $\Delta \bar{a}_{\pi_i}$ indicates the residual difference between the sorted ID and OoD activations. CIFAR-100 and ImageNet-200 exhibit on average weaker high-rank activations than ID, following the structure found in CIFAR-10 (\cref{fig:residuals}). Considering that they share the same model as the CIFAR-10 setup, this suggests that the profile behavior is more model-dependent than dataset-dependent. ConvNeXt-Base and ViT have unrectified penultimate layers and they show strong displacement for both low-rank and high-rank activations, although in opposite directions. The fact that \ours is still well-performing regardless of the missing rectification and of the direction of the displacement is further evidence of its robustness.

\newpage
\section{Complexity}
In~\cref{tab:ood_benchmarking}, we report the computational complexity and mean batch inference time $\bar{t}$ of each score-enhancing method. \ours operates at $\mathcal{O}(D \log D)$ complexity, adding only 6\% overhead over standard EBO inference.

\begin{table}[h!]
\centering
\caption{Efficiency and complexity analysis of score-enhancing post-hoc OoD detection methods. The test was conducted using \SI{10000} inputs fed to an ImageNet-pretrained ResNet18. We timed the average batch forward pass and quantified the additional complexity needed to execute the method.}
\label{tab:ood_benchmarking}
\begin{tabular}{l r r r}
\toprule
\textbf{Method} & \textbf{Complexity} & \quad \textbf{$\bar{t}$ ($\mathrm{ms}/\mathrm{batch}$)} & $\bar{t} / \bar{t}_{EBO}$ \\
\midrule
EBO             & $\mathcal{O}(1)$               & $18.39$ & $-$ \\
\midrule
ReAct           & $\mathcal{O}(D)$               & $18.63$ & $1.01\times$ \\
ASH-P           & $\mathcal{O}(D)$               & $18.82$ & $1.02\times$ \\
ASH-B           & $\mathcal{O}(D)$               & $18.95$ & $1.03\times$ \\
ASH-S           & $\mathcal{O}(D)$               & $19.07$ & $1.04\times$ \\
SCALE           & $\mathcal{O}(D)$               & $19.33$ & $1.05\times$ \\
DICE            & $\mathcal{O}(D N_c)$           & $27.79$ & $1.51\times$ \\
RankFeat        & $\mathcal{O}(KCHW)$            & $46.25$ & $2.51\times$ \\
ODIN            & $\mathcal{O}(\text{Backprop})$  & $89.33$ & $4.86\times$ \\
\rowcolor{gray!20}\textbf{\ours (ours)} & $\mathcal{O}(D \log D)$ & $19.52$ & $1.06\times$ \\
\bottomrule
\multicolumn{4}{p{7.5cm}}{\emph{Notation:} $D$ denotes the penultimate feature dimension, $N_c$ the number of classes, $K$ the number of power iterations, $H \times W$ the spatial resolution of intermediate feature maps, and $C$ the number of channels.}
\end{tabular}
\end{table}

\section{Checkpoints}
\Cref{tab:checkpoints} contains an overview of the sources for the models used in the various paper experiments.

\begin{table}[h]
\centering
\caption{Model checkpoints and evaluation setup used in experiments.}
\label{tab:checkpoints}
\renewcommand{\arraystretch}{1.15}
\resizebox{\linewidth}{!}{
\begin{tabular}{lll @{\hskip 0.5cm} lr}
\toprule
\textbf{Dataset} & \textbf{Model} & \textbf{Model Source} & \textbf{Seed Source} & \textbf{Seeds} \\
\midrule
\multicolumn{5}{l}{\textbf{\ours experiments}} \\
\midrule
CIFAR-10       & ResNet-18       & \texttt{torchvision} & \texttt{OpenOOD} & 3 \\
CIFAR-100      & ResNet-18       & \texttt{torchvision}  & \texttt{OpenOOD} & 3 \\
ImageNet-200   & ResNet-18       & \texttt{torchvision}  & \texttt{OpenOOD} & 3 \\
ImageNet       & ResNet-50       & \texttt{torchvision}   & \texttt{OpenOOD} & 1 \\
ImageNet       & ConvNeXt-Tiny   & \texttt{torchvision} & \texttt{torchvision}         & 1 \\
ImageNet       & ConvNeXt-Base   & \texttt{torchvision} & \texttt{torchvision}         & 1 \\
ImageNet       & EfficientNet-B0 & \texttt{torchvision} & \texttt{torchvision}         & 1 \\
ImageNet       & ViT-B/16        & \texttt{torchvision}   & \texttt{OpenOOD} & 1 \\
ImageNet       & Swin-T          & \texttt{torchvision}   & \texttt{OpenOOD} & 1 \\
\midrule
\multicolumn{5}{l}{\textbf{\ours + MDS/RMDS experiments}} \\
\midrule
CIFAR-10    & ResNet-18
    & \texttt{timm} & \texttt{edadaltocg/resnet18\_cifar10}
    & 1 \\
CIFAR-100   & ResNet-18
    & \texttt{timm} & \texttt{edadaltocg/resnet18\_cifar100}
    & 1 \\
ImageNet    & EfficientNetV2-M
    & \texttt{timm} & \texttt{tf\_efficientnetv2\_m.in1k}
    & 1 \\
ImageNet    & SwinV2-S
    & \texttt{timm} & \texttt{swinv2\_small\_window16\_256.ms\_in1k}
    & 1 \\
ImageNet    & ResNet-50
    & \texttt{timm} & \texttt{resnet50.tv2\_in1k}
    & 1 \\
\bottomrule
\end{tabular}
}
\end{table}

\newpage
\begin{sidewaystable}
\centering
\caption{Extended results for CIFAR-10 on ResNet18.}
\label{tab:complete_1}
\resizebox{\textheight}{!}{
\begin{tabular}{l c c c c c c c c c c c c c c c c c c c c c}
\toprule
 & \multicolumn{3}{c}{\textbf{CIFAR100}} & \multicolumn{3}{c}{\textbf{TIN}} & \multicolumn{3}{c}{\textbf{MNIST}} & \multicolumn{3}{c}{\textbf{SVHN}} & \multicolumn{3}{c}{\textbf{TEXTURE}} & \multicolumn{3}{c}{\textbf{PLACES365}} & \multicolumn{3}{c}{\textbf{AVERAGE}} \\
Method & AUROC & FPR & AUPR & AUROC & FPR & AUPR & AUROC & FPR & AUPR & AUROC & FPR & AUPR & AUROC & FPR & AUPR & AUROC & FPR & AUPR & AUROC & FPR & AUPR \\
\midrule
ASH-B & $74.11 {\scriptstyle \pm 1.90}$ & $87.31 {\scriptstyle \pm 2.53}$ & $77.19 {\scriptstyle \pm 1.95}$ & $76.62 {\scriptstyle \pm 0.74}$ & $85.86 {\scriptstyle \pm 2.03}$ & $77.50 {\scriptstyle \pm 1.23}$ & $83.16 {\scriptstyle \pm 5.71}$ & $70.00 {\scriptstyle \pm 12.94}$ & $97.39 {\scriptstyle \pm 0.90}$ & $73.76 {\scriptstyle \pm 7.67}$ & $83.36 {\scriptstyle \pm 7.79}$ & $89.22 {\scriptstyle \pm 3.76}$ & $77.43 {\scriptstyle \pm 2.93}$ & $84.66 {\scriptstyle \pm 1.98}$ & $72.82 {\scriptstyle \pm 4.62}$ & $79.90 {\scriptstyle \pm 4.53}$ & $77.71 {\scriptstyle \pm 9.23}$ & $94.04 {\scriptstyle \pm 1.24}$ & $77.50 {\scriptstyle \pm 3.91}$ & $81.48 {\scriptstyle \pm 6.08}$ & $84.69 {\scriptstyle \pm 2.28}$ \\
ASH-P & $85.75 {\scriptstyle \pm 0.52}$ & $70.07 {\scriptstyle \pm 4.13}$ & $86.16 {\scriptstyle \pm 0.19}$ & $88.21 {\scriptstyle \pm 0.25}$ & $61.65 {\scriptstyle \pm 5.83}$ & $87.13 {\scriptstyle \pm 0.49}$ & $93.95 {\scriptstyle \pm 3.32}$ & $28.19 {\scriptstyle \pm 19.72}$ & $99.03 {\scriptstyle \pm 0.55}$ & $91.30 {\scriptstyle \pm 1.22}$ & $39.47 {\scriptstyle \pm 8.98}$ & $96.36 {\scriptstyle \pm 0.36}$ & $89.18 {\scriptstyle \pm 0.85}$ & $56.21 {\scriptstyle \pm 7.15}$ & $84.80 {\scriptstyle \pm 0.93}$ & $88.47 {\scriptstyle \pm 1.18}$ & $59.72 {\scriptstyle \pm 7.87}$ & $96.53 {\scriptstyle \pm 0.35}$ & $89.47 {\scriptstyle \pm 1.22}$ & $52.55 {\scriptstyle \pm 8.95}$ & $91.67 {\scriptstyle \pm 0.48}$ \\
\bottomrule
\end{tabular}
}
\end{sidewaystable}

\begin{sidewaystable}
\centering
\caption{Extended results for CIFAR-100 on ResNet18.}
\label{tab:complete_2}
\resizebox{\textheight}{!}{
\begin{tabular}{l c c c c c c c c c c c c c c c c c c c c c}
\toprule
 & \multicolumn{3}{c}{\textbf{CIFAR10}} & \multicolumn{3}{c}{\textbf{TIN}} & \multicolumn{3}{c}{\textbf{MNIST}} & \multicolumn{3}{c}{\textbf{SVHN}} & \multicolumn{3}{c}{\textbf{TEXTURE}} & \multicolumn{3}{c}{\textbf{PLACES365}} & \multicolumn{3}{c}{\textbf{AVERAGE}} \\
Method & AUROC & FPR & AUPR & AUROC & FPR & AUPR & AUROC & FPR & AUPR & AUROC & FPR & AUPR & AUROC & FPR & AUPR & AUROC & FPR & AUPR & AUROC & FPR & AUPR \\
\midrule
ASH-B & $76.48 {\scriptstyle \pm 0.37}$ & $68.06 {\scriptstyle \pm 0.53}$ & $74.93 {\scriptstyle \pm 0.24}$ & $79.79 {\scriptstyle \pm 0.29}$ & $63.47 {\scriptstyle \pm 1.32}$ & $70.30 {\scriptstyle \pm 0.61}$ & $77.23 {\scriptstyle \pm 0.56}$ & $66.58 {\scriptstyle \pm 4.75}$ & $95.45 {\scriptstyle \pm 0.17}$ & $\underline{85.76} {\scriptstyle \pm 1.69}$ & $\underline{45.51} {\scriptstyle \pm 3.45}$ & $\underline{93.04} {\scriptstyle \pm 0.89}$ & $80.72 {\scriptstyle \pm 0.87}$ & $61.25 {\scriptstyle \pm 3.48}$ & $68.98 {\scriptstyle \pm 1.23}$ & $78.76 {\scriptstyle \pm 0.19}$ & $62.88 {\scriptstyle \pm 1.31}$ & $91.47 {\scriptstyle \pm 0.09}$ & $79.79 {\scriptstyle \pm 0.66}$ & $61.29 {\scriptstyle \pm 2.48}$ & $82.36 {\scriptstyle \pm 0.54}$ \\
ASH-P & $\underline{79.22} {\scriptstyle \pm 0.13}$ & $60.64 {\scriptstyle \pm 0.81}$ & $\underline{76.69} {\scriptstyle \pm 0.22}$ & $82.82 {\scriptstyle \pm 0.06}$ & $51.67 {\scriptstyle \pm 0.17}$ & $72.58 {\scriptstyle \pm 0.23}$ & $81.17 {\scriptstyle \pm 1.45}$ & $49.03 {\scriptstyle \pm 5.18}$ & $95.76 {\scriptstyle \pm 0.23}$ & $81.93 {\scriptstyle \pm 1.52}$ & $52.39 {\scriptstyle \pm 4.29}$ & $90.22 {\scriptstyle \pm 1.04}$ & $78.08 {\scriptstyle \pm 1.16}$ & $62.05 {\scriptstyle \pm 2.39}$ & $63.44 {\scriptstyle \pm 1.52}$ & $78.83 {\scriptstyle \pm 0.63}$ & $58.81 {\scriptstyle \pm 1.52}$ & $91.05 {\scriptstyle \pm 0.38}$ & $80.34 {\scriptstyle \pm 0.83}$ & $55.77 {\scriptstyle \pm 2.39}$ & $81.62 {\scriptstyle \pm 0.60}$ \\
\bottomrule
\end{tabular}
}
\end{sidewaystable}

\begin{sidewaystable}
\centering
\caption{Extended results for ImageNet-200 on ResNet18.}
\label{tab:complete_3}
\resizebox{\textheight}{!}{
\begin{tabular}{l c c c c c c c c c c c c c c c c c c}
\toprule
 & \multicolumn{3}{c}{\textbf{SSB\_HARD}} & \multicolumn{3}{c}{\textbf{NINCO}} & \multicolumn{3}{c}{\textbf{INATURALIST}} & \multicolumn{3}{c}{\textbf{TEXTURES}} & \multicolumn{3}{c}{\textbf{OPENIMAGE\_O}} & \multicolumn{3}{c}{\textbf{AVERAGE}} \\
Method & AUROC & FPR & AUPR & AUROC & FPR & AUPR & AUROC & FPR & AUPR & AUROC & FPR & AUPR & AUROC & FPR & AUPR & AUROC & FPR & AUPR \\
\midrule
ASH-B & $79.34 {\scriptstyle \pm 0.41}$ & $72.40 {\scriptstyle \pm 1.22}$ & $94.91 {\scriptstyle \pm 0.10}$ & $85.19 {\scriptstyle \pm 0.09}$ & $57.80 {\scriptstyle \pm 1.42}$ & $\underline{79.02} {\scriptstyle \pm 0.27}$ & $95.08 {\scriptstyle \pm 0.58}$ & $22.53 {\scriptstyle \pm 2.74}$ & $95.41 {\scriptstyle \pm 0.57}$ & $\underline{94.77} {\scriptstyle \pm 0.24}$ & $25.71 {\scriptstyle \pm 0.82}$ & $\underline{92.37} {\scriptstyle \pm 0.49}$ & $\underline{91.80} {\scriptstyle \pm 0.30}$ & $33.96 {\scriptstyle \pm 1.19}$ & $\underline{94.64} {\scriptstyle \pm 0.31}$ & $89.24 {\scriptstyle \pm 0.32}$ & $42.48 {\scriptstyle \pm 1.48}$ & $\underline{91.27} {\scriptstyle \pm 0.35}$ \\
ASH-P & $79.77 {\scriptstyle \pm 0.09}$ & $70.00 {\scriptstyle \pm 0.29}$ & $94.73 {\scriptstyle \pm 0.05}$ & $85.53 {\scriptstyle \pm 0.11}$ & $49.89 {\scriptstyle \pm 0.76}$ & $76.78 {\scriptstyle \pm 0.23}$ & $93.08 {\scriptstyle \pm 0.64}$ & $25.03 {\scriptstyle \pm 2.37}$ & $92.08 {\scriptstyle \pm 0.77}$ & $91.80 {\scriptstyle \pm 0.19}$ & $37.28 {\scriptstyle \pm 2.51}$ & $87.80 {\scriptstyle \pm 0.04}$ & $89.92 {\scriptstyle \pm 0.29}$ & $35.00 {\scriptstyle \pm 1.25}$ & $92.68 {\scriptstyle \pm 0.28}$ & $88.02 {\scriptstyle \pm 0.27}$ & $43.44 {\scriptstyle \pm 1.44}$ & $88.81 {\scriptstyle \pm 0.27}$ \\
\bottomrule
\end{tabular}
}
\end{sidewaystable}

\begin{table}[H]
\centering
\caption{Extended results for ImageNet on ResNet50.}
\label{tab:complete_4}
\resizebox{\linewidth}{!}{
\begin{tabular}{l c c c c c c c c c c c c c c c c c c}
\toprule
 & \multicolumn{3}{c}{\textbf{SSB\_HARD}} & \multicolumn{3}{c}{\textbf{NINCO}} & \multicolumn{3}{c}{\textbf{INATURALIST}} & \multicolumn{3}{c}{\textbf{TEXTURES}} & \multicolumn{3}{c}{\textbf{OPENIMAGE\_O}} & \multicolumn{3}{c}{\textbf{AVERAGE}} \\
Method & AUROC & FPR & AUPR & AUROC & FPR & AUPR & AUROC & FPR & AUPR & AUROC & FPR & AUPR & AUROC & FPR & AUPR & AUROC & FPR & AUPR \\
\midrule
ASH-B & $\underline{73.11}$ & $\underline{73.66}$ & $72.45$ & $\underline{83.37}$ & $52.92$ & $\underline{41.33}$ & $\underline{97.04}$ & $\underline{14.31}$ & $\underline{89.23}$ & $96.91$ & $15.40$ & $84.18$ & $\underline{93.31}$ & $29.23$ & $\underline{84.14}$ & $\underline{88.75}$ & $\underline{37.11}$ & $\underline{74.27}$ \\
ASH-P & $71.90$ & $76.09$ & $70.58$ & $80.66$ & $57.90$ & $32.59$ & $92.43$ & $26.36$ & $70.41$ & $91.56$ & $36.06$ & $64.63$ & $90.67$ & $33.82$ & $75.71$ & $85.44$ & $46.04$ & $62.79$ \\
ASH-S & $\underline{74.59}$ & $\underline{71.43}$ & $\underline{73.93}$ & $\underline{84.41}$ & $53.09$ & $\underline{44.44}$ & $\underline{97.71}$ & $\underline{11.13}$ & $\underline{91.26}$ & $\underline{97.87}$ & $\underline{10.99}$ & $\underline{88.56}$ & $\underline{93.83}$ & $\underline{28.63}$ & $\underline{85.72}$ & $\underline{89.68}$ & $\underline{35.05}$ & $\underline{76.78}$ \\
DICE & $70.13$ & $77.96$ & $69.93$ & $76.01$ & $66.90$ & $31.43$ & $92.54$ & $33.37$ & $77.15$ & $92.04$ & $44.28$ & $73.63$ & $88.26$ & $47.83$ & $75.67$ & $83.80$ & $54.07$ & $65.56$ \\
EBO & $72.08$ & $76.54$ & $70.81$ & $79.70$ & $60.58$ & $30.69$ & $90.63$ & $31.30$ & $64.07$ & $88.70$ & $45.77$ & $55.44$ & $89.06$ & $38.09$ & $71.69$ & $84.03$ & $50.46$ & $58.54$ \\
GEN & $72.01$ & $75.73$ & $69.90$ & $81.70$ & $54.90$ & $31.74$ & $92.44$ & $26.10$ & $71.07$ & $87.59$ & $46.22$ & $45.51$ & $89.26$ & $34.50$ & $69.78$ & $84.60$ & $47.49$ & $57.60$ \\
\bottomrule
\end{tabular}
}
\end{table}

\begin{table}[H]
\centering
\caption{Extended results for ImageNet on ConvNeXt-Tiny.}
\label{tab:complete_5}
\resizebox{\linewidth}{!}{
\begin{tabular}{l c c c c c c c c c c c c c c c c c c}
\toprule
 & \multicolumn{3}{c}{\textbf{SSB\_HARD}} & \multicolumn{3}{c}{\textbf{NINCO}} & \multicolumn{3}{c}{\textbf{INATURALIST}} & \multicolumn{3}{c}{\textbf{TEXTURES}} & \multicolumn{3}{c}{\textbf{OPENIMAGE\_O}} & \multicolumn{3}{c}{\textbf{AVERAGE}} \\
Method & AUROC & FPR & AUPR & AUROC & FPR & AUPR & AUROC & FPR & AUPR & AUROC & FPR & AUPR & AUROC & FPR & AUPR & AUROC & FPR & AUPR \\
\midrule
ASH & $61.89$ & $88.35$ & $60.27$ & $72.24$ & $77.38$ & $21.96$ & $74.43$ & $67.87$ & $31.25$ & $67.43$ & $92.33$ & $17.20$ & $78.31$ & $67.02$ & $50.57$ & $70.86$ & $78.59$ & $36.25$ \\
DICE & $63.64$ & $89.42$ & $\mathbf{66.66}$ & $46.99$ & $97.08$ & $12.11$ & $63.90$ & $90.26$ & $30.14$ & $51.08$ & $96.74$ & $12.44$ & $63.28$ & $95.82$ & $46.50$ & $57.78$ & $93.86$ & $33.57$ \\
EBO & $63.87$ & $87.52$ & $63.61$ & $66.01$ & $86.64$ & $19.77$ & $79.30$ & $66.97$ & $\underline{44.57}$ & $70.12$ & $87.97$ & $30.37$ & $67.49$ & $86.03$ & $41.74$ & $69.36$ & $83.03$ & $40.01$ \\
ReAct & $\underline{67.81}$ & $\underline{82.78}$ & $\underline{65.55}$ & $\underline{75.80}$ & $\underline{74.59}$ & $\underline{26.72}$ & $\underline{83.57}$ & $\underline{50.72}$ & $\underline{46.17}$ & $\underline{81.73}$ & $\underline{73.77}$ & $\mathbf{44.63}$ & $\underline{80.42}$ & $\underline{65.38}$ & $\underline{56.87}$ & $\underline{77.87}$ & $\underline{69.45}$ & $\underline{47.99}$ \\
SCALE & $60.36$ & $86.04$ & $57.94$ & $56.63$ & $86.86$ & $12.60$ & $69.73$ & $71.78$ & $26.82$ & $58.75$ & $88.74$ & $12.61$ & $55.01$ & $88.20$ & $26.72$ & $60.10$ & $84.32$ & $27.34$ \\
\bottomrule
\end{tabular}
}
\end{table}

\begin{table}[H]
\centering
\caption{Extended results for ImageNet on ConvNeXt-Base.}
\label{tab:complete_6}
\resizebox{\linewidth}{!}{
\begin{tabular}{l c c c c c c c c c c c c c c c c c c}
\toprule
 & \multicolumn{3}{c}{\textbf{SSB\_HARD}} & \multicolumn{3}{c}{\textbf{NINCO}} & \multicolumn{3}{c}{\textbf{INATURALIST}} & \multicolumn{3}{c}{\textbf{TEXTURES}} & \multicolumn{3}{c}{\textbf{OPENIMAGE\_O}} & \multicolumn{3}{c}{\textbf{AVERAGE}} \\
Method & AUROC & FPR & AUPR & AUROC & FPR & AUPR & AUROC & FPR & AUPR & AUROC & FPR & AUPR & AUROC & FPR & AUPR & AUROC & FPR & AUPR \\
\midrule
ASH & $32.26$ & $99.44$ & $41.39$ & $29.09$ & $99.75$ & $7.95$ & $4.50$ & $99.97$ & $9.79$ & $42.55$ & $99.51$ & $10.83$ & $22.92$ & $99.92$ & $17.40$ & $26.26$ & $99.72$ & $17.47$ \\
DICE & $66.55$ & $87.00$ & $\underline{68.94}$ & $51.84$ & $97.96$ & $14.89$ & $39.42$ & $99.16$ & $17.43$ & $67.96$ & $96.80$ & $30.75$ & $64.74$ & $95.56$ & $46.51$ & $58.10$ & $95.30$ & $35.70$ \\
EBO & $64.68$ & $90.65$ & $65.80$ & $65.62$ & $91.48$ & $22.23$ & $58.18$ & $91.25$ & $23.71$ & $53.51$ & $97.37$ & $15.88$ & $54.83$ & $95.71$ & $33.19$ & $59.36$ & $93.29$ & $32.16$ \\
ReAct & $\mathbf{72.89}$ & $\mathbf{77.92}$ & $\mathbf{70.33}$ & $\underline{81.44}$ & $\underline{63.09}$ & $\mathbf{32.73}$ & $\underline{88.18}$ & $\underline{35.16}$ & $\underline{53.56}$ & $\underline{78.26}$ & $87.42$ & $\underline{34.54}$ & $\underline{83.49}$ & $\underline{56.72}$ & $\underline{59.79}$ & $\underline{80.85}$ & $\underline{64.06}$ & $\underline{50.19}$ \\
SCALE & $60.87$ & $86.35$ & $59.26$ & $68.27$ & $72.11$ & $18.20$ & $58.47$ & $74.54$ & $19.28$ & $72.29$ & $\underline{76.84}$ & $22.94$ & $66.18$ & $72.79$ & $35.90$ & $65.22$ & $76.53$ & $31.12$ \\
\rowcolor{gray!20} \textbf{\ours} & $\underline{70.63}$ & $\underline{78.56}$ & $67.16$ & $\mathbf{81.98}$ & $\underline{57.31}$ & $\underline{31.18}$ & $\mathbf{91.05}$ & $\mathbf{23.10}$ & $\mathbf{55.49}$ & $\mathbf{89.86}$ & $\mathbf{38.79}$ & $\mathbf{48.50}$ & $\mathbf{90.80}$ & $\mathbf{28.70}$ & $\mathbf{69.37}$ & $\mathbf{84.86}$ & $\mathbf{45.29}$ & $\mathbf{54.34}$ \\
\bottomrule
\end{tabular}
}
\end{table}

\begin{table}[H]
\centering
\caption{Extended results for ImageNet on EfficientNet-B0.}
\label{tab:complete_7}
\resizebox{\linewidth}{!}{
\begin{tabular}{l c c c c c c c c c c c c c c c c c c}
\toprule
 & \multicolumn{3}{c}{\textbf{SSB\_HARD}} & \multicolumn{3}{c}{\textbf{NINCO}} & \multicolumn{3}{c}{\textbf{INATURALIST}} & \multicolumn{3}{c}{\textbf{TEXTURES}} & \multicolumn{3}{c}{\textbf{OPENIMAGE\_O}} & \multicolumn{3}{c}{\textbf{AVERAGE}} \\
Method & AUROC & FPR & AUPR & AUROC & FPR & AUPR & AUROC & FPR & AUPR & AUROC & FPR & AUPR & AUROC & FPR & AUPR & AUROC & FPR & AUPR \\
\midrule
ASH & $54.01$ & $87.63$ & $51.30$ & $50.62$ & $86.30$ & $10.52$ & $53.65$ & $80.07$ & $17.41$ & $67.01$ & $70.44$ & $13.91$ & $54.64$ & $83.91$ & $25.61$ & $55.99$ & $81.67$ & $23.75$ \\
EBO & $\mathbf{71.75}$ & $\mathbf{81.48}$ & $\mathbf{72.70}$ & $75.08$ & $76.93$ & $27.27$ & $79.79$ & $61.86$ & $41.05$ & $77.34$ & $74.24$ & $28.46$ & $79.85$ & $66.95$ & $55.71$ & $76.76$ & $72.29$ & $45.04$ \\
DICE & $47.77$ & $90.96$ & $48.21$ & $37.62$ & $94.38$ & $8.43$ & $36.47$ & $91.82$ & $13.24$ & $55.99$ & $85.29$ & $11.32$ & $38.74$ & $94.22$ & $20.18$ & $43.32$ & $91.33$ & $20.28$ \\
ReAct & $68.22$ & $84.44$ & $\underline{67.52}$ & $\underline{79.89}$ & $\underline{61.83}$ & $\underline{31.50}$ & $\mathbf{89.83}$ & $\underline{34.32}$ & $\mathbf{61.17}$ & $\mathbf{93.51}$ & $\mathbf{28.84}$ & $\mathbf{66.05}$ & $\mathbf{89.90}$ & $\underline{35.60}$ & $\mathbf{72.56}$ & $\mathbf{84.27}$ & $\underline{49.01}$ & $\mathbf{59.76}$ \\
SCALE & $54.82$ & $87.50$ & $51.89$ & $52.94$ & $85.36$ & $11.04$ & $57.86$ & $77.58$ & $18.98$ & $71.21$ & $66.37$ & $16.11$ & $58.40$ & $81.68$ & $27.63$ & $59.05$ & $79.70$ & $25.13$ \\
\rowcolor{gray!20} \textbf{\ours } & $\underline{68.25}$ & $\underline{83.20}$ & $\underline{66.61}$ & $\underline{81.56}$ & $\underline{58.00}$ & $\underline{31.90}$ & $\underline{89.44}$ & $\mathbf{27.22}$ & $\underline{54.70}$ & $\underline{90.25}$ & $\underline{34.76}$ & $\underline{51.74}$ & $\underline{89.42}$ & $\mathbf{31.16}$ & $\underline{69.10}$ & $\underline{83.78}$ & $\mathbf{46.87}$ & $\underline{54.81}$ \\
\bottomrule
\end{tabular}
}
\end{table}

\begin{table}[H]
\centering
\caption{Extended results for ImageNet on ViT.}
\label{tab:complete_8}
\resizebox{\linewidth}{!}{
\begin{tabular}{l c c c c c c c c c c c c c c c c c c}
\toprule
 & \multicolumn{3}{c}{\textbf{SSB\_HARD}} & \multicolumn{3}{c}{\textbf{NINCO}} & \multicolumn{3}{c}{\textbf{INATURALIST}} & \multicolumn{3}{c}{\textbf{TEXTURES}} & \multicolumn{3}{c}{\textbf{OPENIMAGE\_O}} & \multicolumn{3}{c}{\textbf{AVERAGE}} \\
Method & AUROC & FPR & AUPR & AUROC & FPR & AUPR & AUROC & FPR & AUPR & AUROC & FPR & AUPR & AUROC & FPR & AUPR & AUROC & FPR & AUPR \\
\midrule
OpenMax & $68.60$ & $89.19$ & $\underline{72.54}$ & $78.68$ & $88.33$ & $41.68$ & $94.93$ & $\underline{19.62}$ & $81.73$ & $85.52$ & $73.07$ & $46.39$ & $87.36$ & $73.82$ & $73.54$ & $83.02$ & $68.81$ & $63.18$ \\
MSP & $68.94$ & $86.41$ & $68.58$ & $78.11$ & $77.28$ & $32.03$ & $88.19$ & $42.40$ & $63.35$ & $85.06$ & $56.46$ & $44.28$ & $84.86$ & $56.19$ & $66.27$ & $81.03$ & $63.75$ & $54.90$ \\
TempScale & $68.55$ & $87.35$ & $68.49$ & $77.80$ & $81.88$ & $32.35$ & $88.54$ & $43.09$ & $64.38$ & $85.39$ & $58.16$ & $45.49$ & $85.04$ & $59.98$ & $67.07$ & $81.06$ & $66.09$ & $55.56$ \\
MDS & $\underline{71.57}$ & $\underline{83.47}$ & $\underline{73.88}$ & $\underline{86.52}$ & $\underline{48.77}$ & $\underline{49.42}$ & $\underline{96.01}$ & $20.64$ & $\mathbf{86.08}$ & $89.41$ & $38.91$ & $49.10$ & $\mathbf{92.38}$ & $\underline{30.35}$ & $\mathbf{79.68}$ & $\underline{87.18}$ & $\underline{44.43}$ & $\underline{67.63}$ \\
RMDS & $\mathbf{72.87}$ & $\underline{84.52}$ & $\mathbf{75.09}$ & $\mathbf{87.31}$ & $\mathbf{46.20}$ & $\mathbf{50.20}$ & $\mathbf{96.10}$ & $\underline{19.47}$ & $\underline{86.00}$ & $89.38$ & $\underline{37.22}$ & $46.01$ & $\underline{92.32}$ & $\mathbf{29.57}$ & $\underline{79.31}$ & $\mathbf{87.60}$ & $\mathbf{43.40}$ & $\underline{67.32}$ \\
EBO & $58.80$ & $92.24$ & $61.41$ & $66.02$ & $94.14$ & $24.76$ & $79.30$ & $83.56$ & $49.87$ & $81.17$ & $83.66$ & $43.98$ & $76.48$ & $88.82$ & $58.81$ & $72.35$ & $88.48$ & $47.77$ \\
\bottomrule
\end{tabular}
}
\end{table}

\begin{table}[H]
\centering
\caption{Extended results for ImageNet on Swin-T.}
\label{tab:complete_9}
\resizebox{\linewidth}{!}{
\begin{tabular}{l c c c c c c c c c c c c c c c c c c}
\toprule
 & \multicolumn{3}{c}{\textbf{SSB\_HARD}} & \multicolumn{3}{c}{\textbf{NINCO}} & \multicolumn{3}{c}{\textbf{INATURALIST}} & \multicolumn{3}{c}{\textbf{TEXTURES}} & \multicolumn{3}{c}{\textbf{OPENIMAGE\_O}} & \multicolumn{3}{c}{\textbf{AVERAGE}} \\
Method & AUROC & FPR & AUPR & AUROC & FPR & AUPR & AUROC & FPR & AUPR & AUROC & FPR & AUPR & AUROC & FPR & AUPR & AUROC & FPR & AUPR \\
\midrule
OpenMax & $71.56$ & $85.50$ & $\mathbf{72.73}$ & $81.66$ & $72.58$ & $37.16$ & $\underline{95.08}$ & $\underline{19.28}$ & $\underline{79.45}$ & $82.75$ & $77.31$ & $39.13$ & $87.84$ & $60.76$ & $71.94$ & $83.78$ & $63.09$ & $60.08$ \\
MSP & $\underline{71.78}$ & $\underline{80.90}$ & $70.91$ & $81.72$ & $61.21$ & $35.73$ & $89.86$ & $37.31$ & $66.40$ & $83.27$ & $60.87$ & $41.31$ & $85.77$ & $49.30$ & $66.84$ & $82.48$ & $57.92$ & $56.24$ \\
TempScale & $\underline{71.86}$ & $\underline{82.25}$ & $71.17$ & $82.10$ & $63.24$ & $36.60$ & $90.37$ & $36.66$ & $67.56$ & $83.68$ & $62.55$ & $42.54$ & $86.14$ & $50.93$ & $67.78$ & $82.83$ & $59.13$ & $57.13$ \\
ODIN & $66.94$ & $86.45$ & $66.51$ & $69.35$ & $86.05$ & $23.00$ & $78.10$ & $79.35$ & $49.39$ & $72.86$ & $85.46$ & $31.67$ & $68.93$ & $89.82$ & $47.94$ & $71.24$ & $85.43$ & $43.70$ \\
MDS & $68.61$ & $83.60$ & $67.43$ & $81.75$ & $\underline{53.71}$ & $32.03$ & $93.59$ & $21.56$ & $73.69$ & $\underline{89.81}$ & $\underline{37.49}$ & $52.94$ & $\underline{91.05}$ & $\underline{30.56}$ & $\underline{74.56}$ & $84.96$ & $45.38$ & $60.13$ \\
RMDS & $71.77$ & $82.77$ & $\underline{72.04}$ & $\underline{84.92}$ & $\underline{50.13}$ & $\underline{40.44}$ & $\mathbf{95.58}$ & $\underline{18.44}$ & $\mathbf{83.33}$ & $89.26$ & $39.18$ & $52.09$ & $\underline{92.17}$ & $\underline{28.91}$ & $\mathbf{78.93}$ & $\mathbf{86.74}$ & $\mathbf{43.89}$ & $\mathbf{65.37}$ \\
\bottomrule
\end{tabular}
}
\end{table}


\end{document}